\documentclass[final]{cvpr}

\usepackage{times}
\usepackage{epsfig}
\usepackage{graphicx}
\usepackage{amsmath}
\usepackage{amssymb}

\usepackage[font=small,labelfont=bf]{caption}
\usepackage{multirow}
\usepackage{footmisc}
\usepackage{bm}
\usepackage{transparent}

\usepackage{algorithm}
\usepackage{algorithmic}
\usepackage{bbm}

\newcommand*\inlineimage[1]{\raisebox{-0.14\baselineskip}{\includegraphics[height=0.8\baselineskip]{#1}}}
\newcommand{\csobjecta}{\inlineimage{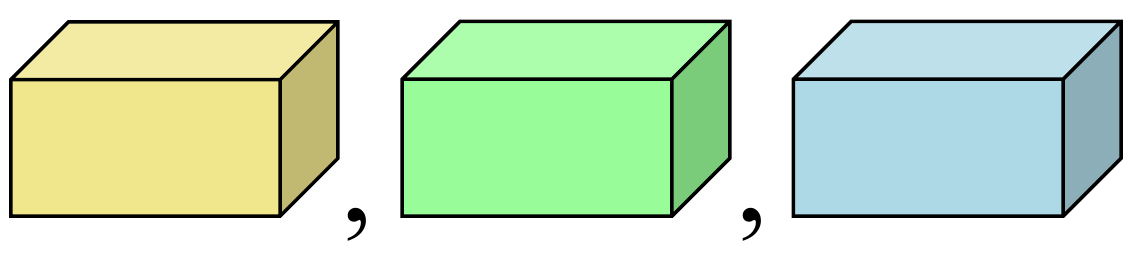}}
\newcommand{\csobjectb}{\inlineimage{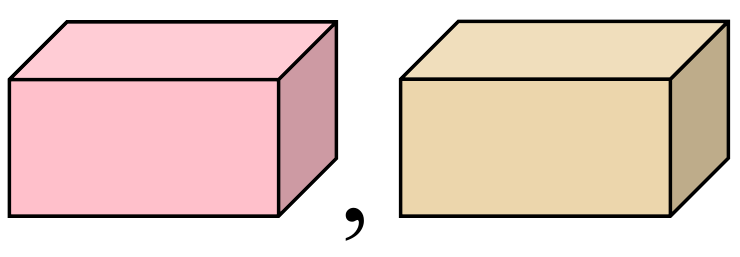}}

\usepackage{colortbl} 
\definecolor{mygray}{gray}{0.6}

\newcommand\blfootnote[1]{%
	\begingroup
	\renewcommand\thefootnote{}\footnote{#1}%
	\addtocounter{footnote}{-1}%
	\endgroup
}

\usepackage[pagebackref=true,breaklinks=true,colorlinks,bookmarks=false]{hyperref}

\def\cvprPaperID{6940} 
\def\confYear{CVPR 2021}

\begin{document}
	
\title{Human-like Controllable Image Captioning with Verb-specific Semantic Roles}

\author{Long Chen$^{2,3\ast}$ \qquad Zhihong Jiang$^{1\ast}$ \qquad Jun Xiao$^{1\dagger}$ \qquad Wei Liu$^4$ \\
	$^1$Zhejiang University \quad $^2$Tencent AI Lab \quad $^3$Columbia University \quad $^4$Tencent Data Platform \\
	{\tt\small zjuchenlong@gmail.com, \{zju\_jiangzhihong, junx\}@zju.edu.cn, wl2223@columbia.edu}
}



\thispagestyle{empty}
\twocolumn[{%
	\maketitle
	\thispagestyle{empty}
	\begin{center}
		\centering
		\includegraphics[width=0.97\linewidth]{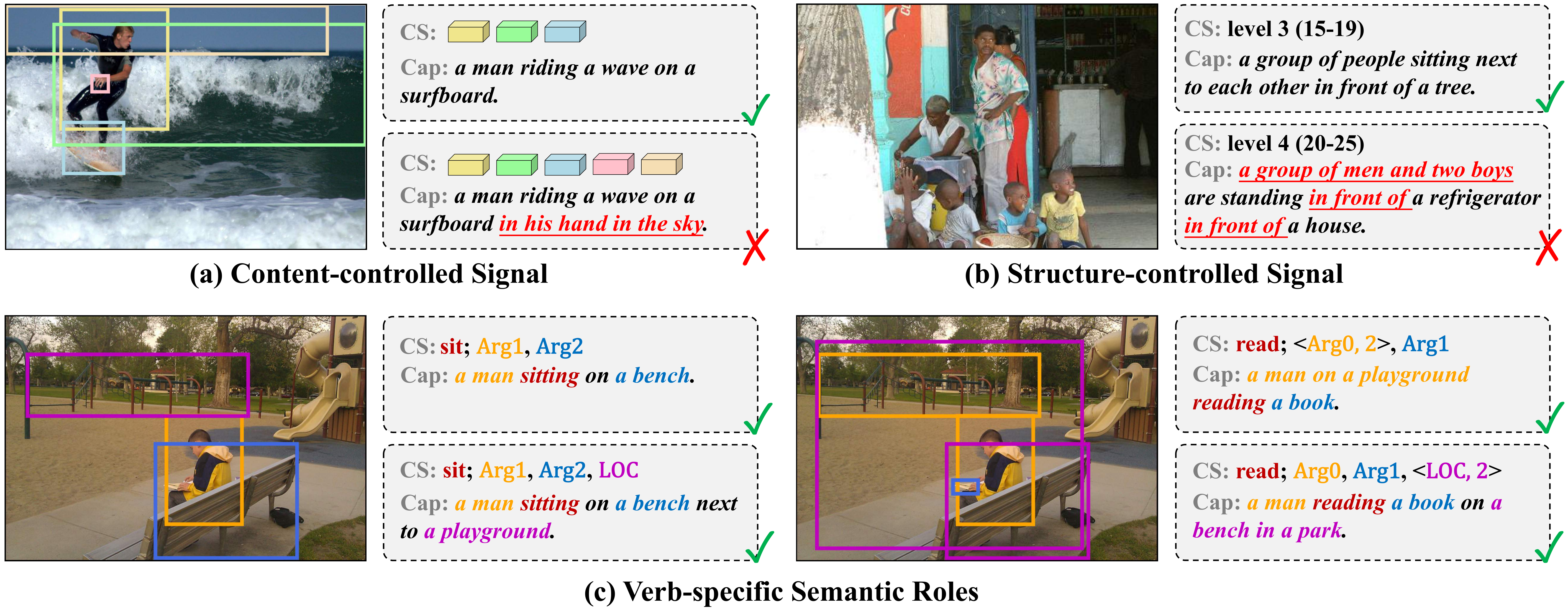}	
		\vspace{-0.8em}
		\captionof{figure}{The CS and Cap in each sample denote the \underline{c}ontrol \underline{s}ignal and generated \underline{cap}tion, respectively. \textbf{(a):} The captions are generated by model SCT~\cite{cornia2019show}, which uses a set of visual regions as control signals. When control signals don't meet the \textbf{event-compatible} requirement (\eg, objects \texttt{hand} and \texttt{sky}), SCT generates lower quality captions (red cross). \textbf{(b):} The captions are generated by model LaBERT~\cite{deng2020length}, which uses different length-levels as control signals. When control signals don't meet the \textbf{sample-suitable} requirement (\eg, level 4), the LaBERT generates lower quality captions. \textbf{(c)}: The captions are generated by our framework with VSR as control signals. For brevity, we abbreviate $<\texttt{role}, \texttt{1}>$ to \texttt{role} in all samples. \texttt{Arg} and \texttt{LOC} denote ``argument" and ``location", respectively. For verb \texttt{sit}, \texttt{Arg1} and \texttt{Arg2} are ``thing sitting" and ``sitting position", respectively. For verb \texttt{read}, \texttt{Arg0} and \texttt{Arg1} are ``reader" and ``thing read", respectively.
		}
		\label{fig:motivation}
	\end{center}%
}]

\begin{abstract}
	
	\blfootnote{$^\ast$ denotes equal contributions, $^\dagger$ denotes the corresponding author.}Controllable Image Captioning (CIC) --- generating image descriptions following designated control signals --- has received unprecedented attention over the last few years. To emulate the human ability in controlling caption generation, current CIC studies focus exclusively on control signals concerning objective properties, such as contents of interest or descriptive patterns. However, we argue that almost all existing objective control signals have overlooked two indispensable characteristics of an ideal control signal: 1) Event-compatible: all visual contents referred to in a single sentence should be compatible with the described activity. 2) Sample-suitable: the control signals should be suitable for a specific image sample. To this end, we propose a new control signal for CIC: Verb-specific Semantic Roles (VSR). VSR consists of a verb and some semantic roles, which represents a targeted activity and the roles of entities involved in this activity. Given a designated VSR, we first train a grounded semantic role labeling (GSRL) model to identify and ground all entities for each role. Then, we propose a semantic structure planner (SSP) to learn human-like descriptive semantic structures. Lastly, we use a role-shift captioning model to generate the captions. Extensive experiments and ablations demonstrate that our framework can achieve better controllability than several strong baselines on two challenging CIC benchmarks. Besides, we can generate multi-level diverse captions easily. The code is available at: \href{https://github.com/mad-red/VSR-guided-CIC}{https://github.com/mad-red/VSR-guided-CIC}.

	
\end{abstract}

\section{Introduction}

Image captioning, \ie, generating fluent and meaningful descriptions to summarize the salient contents of an image, is a classic proxy task for comprehensive scene understanding~\cite{farhadi2010every}. With the release of several large scale datasets and advanced encoder-decoder frameworks, current captioning models plausibly have already achieved ``super-human" performance in all accuracy-based evaluation metrics. However, many studies have indicated that these models tend to produce generic descriptions, and fail to control the caption generation process as humans, \eg, referring to different contents of interest or descriptive patterns. In order to endow the captioning models with human-like controllability, a recent surge of efforts~\cite{cornia2019show,chen2020say,deng2020length,zhong2020comprehensive,pont2020connecting,zheng2019intention,kim2019dense,deshpande2019fast} resort to introducing extra control signals as constraints of the generated captions, called \textbf{Controllable Image Captioning} (CIC). As a byproduct, the CIC models can easily generate diverse descriptions by feeding different control signals.

Early CIC works mainly focus on \textbf{\emph{subjective}} control signals, such as sentiments~\cite{mathews2016senticap}, emotions~\cite{mathews2018semstyle,gan2017stylenet}, and personality~\cite{chunseong2017attend,shuster2019engaging}, \ie, the linguistic styles of sentences. Although these stylized captioning models can eventually produce style-related captions, they remain hard to control the generation process effectively and precisely. To further improve the controllability, recent CIC works gradually put a more emphasis on \textbf{\emph{objective}} control signals. More specifically, they can be coarsely classified into two categories: 1) \emph{Content-controlled}: the control signals are about the contents of interest which need to be described. As the example shown in Figure~\ref{fig:motivation} (a), given the region set (\csobjecta) as a control signal, we hope that the generated caption can cover all regions (\ie, \texttt{man}, \texttt{wave}, and \texttt{surfboard}). So far, various types of content-controlled signals have been proposed, such as visual relations~\cite{kim2019dense}, object regions~\cite{cornia2019show,lindh2020language}, scene graphs~\cite{chen2020say,zhong2020comprehensive}, and mouse trace~\cite{pont2020connecting}. 2) \emph{Structure-controlled}: the control signals are about the semantic structures of sentences. For instance, the length-level~\cite{deng2020length}, part-of-speech tags~\cite{deshpande2019fast}, or attributes~\cite{zhu2021macroscopic} of the sentence (cf. Figure~\ref{fig:motivation} (b)) are some typical structure-controlled signals.

Nevertheless, all existing objective control signals (\ie, both content-controlled and structure-controlled) have overlooked two indispensable characteristics of an ideal control signal towards ``human-like" controllable image captioning: 1) \textbf{Event-compatible}: all visual contents referred to in a single sentence should be compatible with the described activity. Imaging how humans describe images --- our brains always quickly structure a descriptive pattern like ``\textsc{sth do sth at someplace}" first, and then fill in the detailed description~\cite{slevc2011saying,pickering2013integrated,lee2013ways,yang2019learning}, \ie, we have subconsciously made sure that all the mentioned entities are event-compatible (\eg, \texttt{man}, \texttt{wave}, \texttt{surfboard} are all involved in activity \texttt{riding} in Figure~\ref{fig:motivation} (a)). To further see the negative impact of dissatisfying this requirement, suppose that we deliberately utilize two more objects (\texttt{hand} and \texttt{sky}, \ie, \csobjectb) as part of the control signal, and the model generates an incoherent and illogical caption. 2) \textbf{Sample-suitable}: the control signals should be suitable for the specific image sample. By ``suitable", we mean that there do exist reasonable descriptions satisfying the control signals, \eg, a large length-level may not be suitable for an image with a very simple scene. Unfortunately, it is always very difficult to decide whether a control signal is sample-suitable in advance. For example in Figure~\ref{fig:motivation} (b), although the two control signals (\ie, length-levels 3 and 4) are quite close, the quality of respectively generated captions varies greatly.

\begin{figure}[t]
	\begin{minipage}[c]{0.63\linewidth}
		\includegraphics[width=1.0\linewidth]{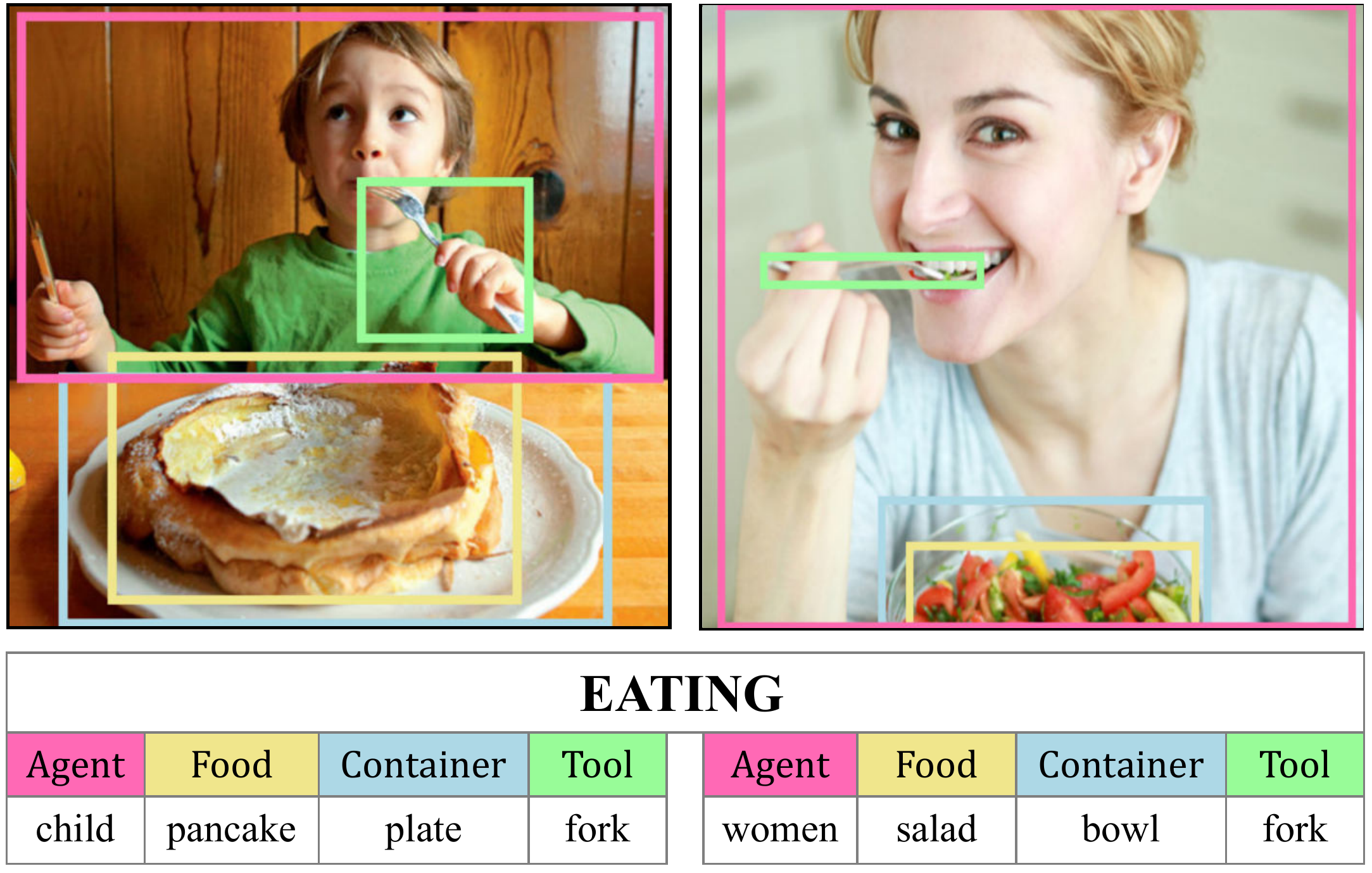}
	\end{minipage}\hfill
	\begin{minipage}[c]{0.33\linewidth}
		\caption{\footnotesize Two image examples of a verb and its semantic roles. The verb \texttt{eating} captures the scope of the activity, and \texttt{agent}, \texttt{food}, \texttt{container}, \texttt{tool} are all reasonable semantic roles for this activity.}
		\label{fig:VSR}
	\end{minipage}
	\vspace{-1.0em}
\end{figure}


In this paper, we propose a new event-oriented objective control signal, \emph{Verb-specific Semantic Roles} (VSR), to meet both event-compatible and sample-suitable requirements simultaneously. VSR consists of a verb (\ie, predicate~\cite{chen2019counterfactual}) and some user-interested semantic roles~\cite{levin1993english}. As shown in Figure~\ref{fig:VSR}, the verb captures the scope of a salient activity in the image (\eg, \texttt{eating}), and the corresponding semantic roles\footnote{We use PropBank-style annotations of semantic roles (\eg, \texttt{Arg0}, \texttt{Arg1}) in all experiments (cf. Figure~\ref{fig:motivation}). The FrameNet-style annotations of semantic roles (\eg, \texttt{Agent}) here are just for a more intuitive illustration. In the PropBank-style annotations, \texttt{Arg} denotes ``argument", $\texttt{MNR}$ denotes ``manner", $\texttt{DIR}$ denotes ``directional", and $\texttt{LOC}$ denotes ``location". We leave more details in the supplementary material. \label{ftnote:sr}} (\eg, \texttt{agent}, \texttt{food}, \texttt{container}, and \texttt{tool}) categorize how objects participate in this activity, \ie, a child (\texttt{agent}) is eating (\texttt{activity}) a pancake (\texttt{food}) from a plate (\texttt{container}) with a fork (\texttt{tool}). Thus, VSR is designed to guarantee that all the mentioned entities are event-compatible. Meanwhile, unlike the existing structure-controlled signals which directly impose constraints on the generated captions, VSR only restricts the involved semantic roles, which is theoretically suitable for all the images with the activity, \ie, sample-suitable.

In order to generate sentences with respect to the designated VSRs, we first train a grounded semantic role labeling (GSRL) model to identify and ground all entities for each role. Then, we propose a semantic structure planner (SSP) to rank the given verb and semantic roles, and output some human-like descriptive semantic structures, \eg, \texttt{Arg0}$_{\text{reader}}$ --  \texttt{read} -- \texttt{Arg1}$_{\text{thing}}$ -- \texttt{LOC} in Figure~\ref{fig:motivation} (c). Finally, we combine the grounded entities and semantic structures, and use an RNN-based role-shift captioning model to generate the captions by sequentially focusing on different roles.

Although these are no available captioning datasets with the VSR annotations, they can be easily obtained by off-the-shelf semantic role parsing toolkits~\cite{shi2019simple}. Extensive experiments on two challenging CIC benchmarks (\ie, COCO Entities~\cite{cornia2019show} and Flickr30K Entities~\cite{plummer2015flickr30k}) demonstrate that our framework can achieve better controllability given designated VSRs than several strong baselines. Moreover, our framework can also realize diverse image captioning and achieve a better trade-off between quality and diversity.

In summary, we make three contributions in this paper:
\vspace{-0.6em}
\begin{enumerate}
	\itemsep-0.4em
	
	\item We propose a new control signal for CIC: Verb-specific Semantic Roles (VSR). To the best of our knowledge, VSR is the first control signal to consider both event-compatible and sample-suitable requirements\footnote{When using control signals extracted from GT captions, existing control signals can always meet both requirements and generate reasonable captions. However, in more general settings (\eg, construct control signals without GT captions), the form of VSR is more human-friendly, and it is easier to construct signals which meet both requirements compared with all existing forms of control signals, which is the main advantage of VSR.}.
	
	\item We can learn human-like verb-specific semantic structures automatically, and abundant visualization examples demonstrate that these patterns are reasonable.
	
	\item We achieve state-of-the-art controllability on two challenging benchmarks, and generate diverse captions by using different verbs, semantic roles, or structures.
	
\end{enumerate}

\section{Related Work}


\noindent\textbf{Controllable Image Captioning.}
Compared with conventional image captioning~\cite{vinyals2015show,xu2015show,chen2017sca,jiang2018recurrent,chen2018regularizing}, CIC is a more challenging task, which needs to consider extra constraints. Early CIC works are mostly about stylized image captioning, \ie, constraints are the linguistic styles of sentences. According to the requirements of parallel training samples, existing solutions can be divided into two types: models using parallel stylized image-caption data~\cite{mathews2016senticap,chen2018factual,shuster2019engaging,alikhani2020clue} or not~\cite{gan2017stylenet,mathews2018semstyle}. Subsequently, the community gradually shifts the emphasis to controlling described contents~\cite{cornia2019show,zheng2019intention,kim2019dense,chen2020say,zhong2020comprehensive,pont2020connecting,lindh2020language} or structures~\cite{deshpande2019fast,deng2020length,yuan2020controllable,zheng2020syntax} of the sentences. In this paper, we propose a novel control signal VSR, which is the first control signal to consider both the event-compatible and sample-suitable requirements.

\noindent\textbf{Diverse and Distinctive Image Captioning.}
Diverse image captioning, \ie, describing the image contents with diverse wordings and rich expressions, is an essential property of human-like captioning models. Except from feeding different control signals to the CIC models, other diverse captioning methods can be coarsely grouped into four types:
1) GAN-based~\cite{dai2017towards,shetty2017speaking,li2018generating}: they use a discriminator to force the generator to generate human-indistinguishable captions. 2) VAE-based~\cite{wang2017diverse,chen2019variational}: the diversity obtained with them is by sampling from a learned latent space. 3) RL-based~\cite{luo2018discriminability}: they regard diversity as an extra reward in the RL training stage. 4) BS-based~\cite{vijayakumar2018diverse}: they decode a list of diverse captions by optimizing a diversity-augmented objective.

Meanwhile, distinctive image captioning is another close research direction~\cite{dai2017contrastive,vedantam2017context,liu2018show,liu2019generating,wang2020compare}, which aims to generate discriminative and unique captions for individual images. Unfortunately, due to the subjective nature of diverse and distinctive captions, effective evaluation remains as an open problem, and several new metrics are proposed, such as SPICE-U~\cite{wang2020towards}, CIDErBtw~\cite{wang2020compare}, self-CIDEr~\cite{wang2019describing}, word recall~\cite{van2018measuring}, mBLEU~\cite{shetty2017speaking}. In this paper, we can easily generate diverse captions in both lexical-level and syntactic-level.

\noindent\textbf{Semantic Roles in Images.} Inspired from the semantic role labeling task~\cite{carreras2005introduction} in NLP, several tasks have been proposed to label the roles of each object in an activity in an image:

\emph{Visual Semantic Role Labeling (VSRL)}, also called situation recognition, is a generalization of action recognition and human-object interaction, which aims to label an image with a set of verb-specific action \emph{frames}~\cite{yatskar2016situation}. Specifically, each action frame describes details of the activity captured by the verb, and it consists of a fixed set of verb-specific semantic roles and their corresponding values. The values are the entities or objects involved in the activity and the semantic roles categorize how objects participate in the activity. The current VSRL methods~\cite{gupta2015visual,yatskar2016situation,mallya2017recurrent,li2017situation,yatskar2017commonly,suhail2019mixture,cooray2020attention} usually learn an independent action classifier first, and then model the role inter-dependency by RNNs or GNNs.

\emph{Grounded Semantic Role Labeling (GSRL)}, also called grounded situation recognition, builds upon the VSRL task, which requires the models not only to label a set of frames, but also to localize each role-value pair in the image~\cite{pratt2020grounded,silberer2018grounding,yang2016grounded,gupta2015visual}. 
In this paper, we use the GSRL model as a bridge to connect the control signals (VSR) and related regions. To the best of our knowledge, we are the first captioning work to benefit from the verb lexicon developed by linguists.

\begin{figure*}[t]
	\centering
	\includegraphics[width=0.98\linewidth]{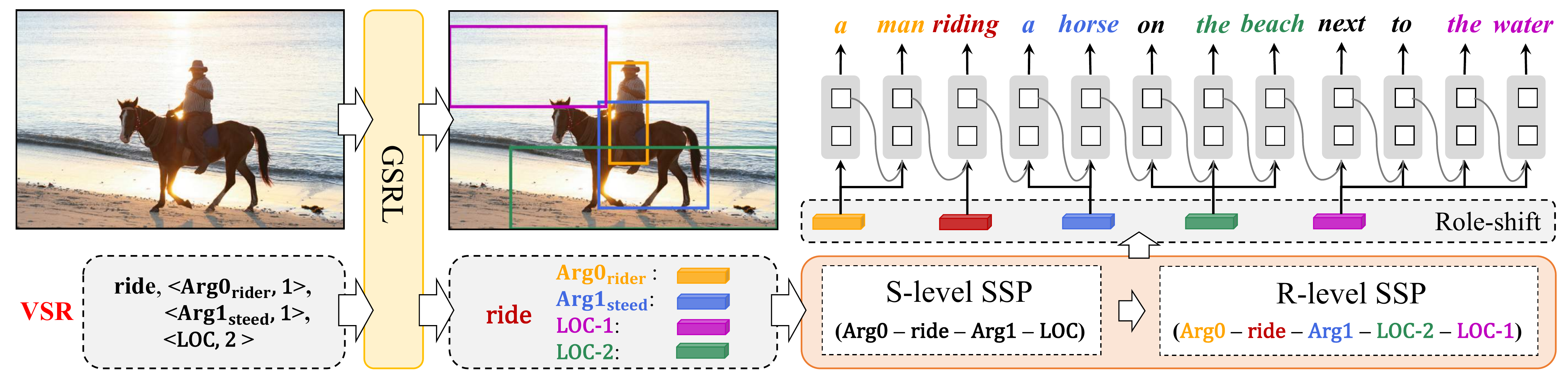}
	\vspace{-0.8em}
	\caption{The whole architecture of our proposed VSR-guided CIC model. This framework consists of three components: 1) a GSRL model to ground the entities for each role; 2) an SSP to learn a semantic structure; 3) a role-shift captioning model to generate the caption.}
	\vspace{-1.0em}
	\label{fig:architecture}
\end{figure*}

\section{Proposed Approach}

For human-like controllable image captioning, we first propose the Verb-specific Semantic Roles (VSR) as the control signal for generating customized captions. As shown in Figure~\ref{fig:architecture}, we formally represent a control signal VSR as:
\begin{eqnarray}
\begin{aligned}
\mathcal{VSR} = \{v, <s_1, n_1>, ..., <s_m, n_m>\}, \\
\end{aligned}
\end{eqnarray}
where $v$ is a \textbf{verb} capturing the scope of a salient activity in the image (\eg, \texttt{ride}), $s_i$ is a \textbf{semantic role} of verb $v$ (\eg, \texttt{LOC}), and $n_i$ is the number of interested entities in the role $s_i$. For example, for $\mathcal{VSR} = \{\texttt{ride}, <\texttt{Arg0}, \texttt{1}>, <\texttt{Arg1}, \texttt{1}>, <\texttt{Loc}, \texttt{2}> \}$, we hope to generate a caption which not only focuses on describing the \texttt{ride} activity, but also contains one entity respectively in the role \texttt{Arg0}$_\text{rider}$ and \texttt{Arg1}$_\text{steed}$, and two entities in the role \texttt{LOC}. Thus, VSR can effectively control the amount of information carried in the whole sentence and each role, \ie, the level of details.

It is convenient to construct VSRs automatically or manually. For the verbs, they can be accurately predicted by an off-the-shelf action recognition network with a predefined verb vocabulary. For the verb-specific semantic roles, they can be easily retrieved from the verb lexicon such as PropBank or FrameNet. Then, the users can easily select a subset of roles or an automatic sampling to generate a subset of roles, and randomly assign the entity number for each role.

Given an image $\bm{I}$ and a control signal $\mathcal{VSR}$, the controllable image captioning model aims to describe $\bm{I}$ by a textual sentence $\bm{y} = \{y_1, ..., y_T\}$, \ie, modeling the probability $p(\bm{y} | \bm{I}, \mathcal{VSR})$. Inspired from the human habit of describing images, we decompose this task into two steps: structuring a descriptive pattern and filling in detailed captions:
\begin{equation} \label{eq:2}
\begin{aligned}
p(\bm{y} | \bm{I}, \mathcal{VSR}) = p(\bm{y}| \text{pattern} ) p(\text{pattern} | \bm{I}, \mathcal{VSR}).
\end{aligned}
\end{equation}

Further, we utilize two sequences $\mathcal{S} = (s^b_1, ..., s^b_K)$ and $\mathcal{R} = (\bm{r}_1, ..., \bm{r}_K)$  to model the descriptive patterns. Specifically, $\mathcal{S}$ is a semantic structure of the sentence and each $s^b_i \in \mathcal{S}$ is a sub-role. By ``sub-role", we mean that each role $s_i \in \mathcal{VSR}$ can be divided into $n_i$ sub-roles, and when $n_i =1$, role $s_i$ itself is a sub-role. Thus, VSR in Figure~\ref{fig:architecture} can be rewritten as \texttt{Arg0}, \texttt{Arg1}, \texttt{LOC-1}, and \texttt{LOC-2}. $\mathcal{R}$ is a sequence of visual features of the corresponding grounded entities for each sub-role in $\mathcal{S}$ (\eg, $\bm{r}_i$ is the features of visual regions referring to $s^b_i$). Particularly, for presentation conciseness, we regard the verb in $\mathcal{VSR}$ as a special type of sub-role, and since there are no grounded visual regions referring to the verb, we use the global image feature as the grounded region feature in $\mathcal{R}$. Meanwhile, we use $\mathcal{\tilde{R}}$ to denote a set of all elements in the sequence $\mathcal{R}$. Thus, we further decompose this task into three components:
\begin{equation} \label{eq:3}
\begin{aligned}
p(\bm{y} | \bm{I}, \mathcal{VSR})  = \underbrace{p(\bm{y} | \mathcal{S}, \mathcal{R})}_{\text{Captioner}} \underbrace{p(\mathcal{S}, \mathcal{R} | \mathcal{\tilde{R}}, \mathcal{VSR})}_{\text{SSP}} \underbrace{p(\mathcal{\tilde{R}} | \bm{I}, \mathcal{VSR} )}_{\text{GSRL}}.    \\
\end{aligned}
\end{equation}

In this section, we first introduce each component of the whole framework of the VSR-guided controllable image captioning model sequentially in Section~\ref{sec:wholeVSR} (cf. Figure~\ref{fig:architecture}), including a grounded semantic role labeling (GSRL) model, a semantic structure planner (SSP), and a role-shift captioning model. Then, we demonstrate the details about all training objectives and the inference stage in Section~\ref{sec:traintest}, including extending from a single VSR to multiple VSRs.

\subsection{Controllable Caption Generation with VSR} \label{sec:wholeVSR}

\subsubsection{Grounded Semantic Role Labeling (GSRL)}

Given an image $\bm{I}$, we first utilize an object detector~\cite{ren2016faster} to extract a set of object proposals $\mathcal{B}$. Each proposal $\bm{b}_i \in \mathcal{B}$ is associated with a visual feature $\bm{f}_i$ and a class label $c_i \in \mathcal{C}$. Then, we group all these proposals into $N$ disjoint sets, \ie, $\mathcal{B} = \{\mathcal{B}_1, ..., \mathcal{B}_N\}$\footnote{Due to different annotation natures of specific CIC datasets, we group proposals by different principles. Details are shown in Section~\ref{sec:details}.}, and each proposal set $\mathcal{B}_i$ consists of one or more proposals. In this GSRL step, we  need to refer each sub-role in the $\mathcal{VSR}$ to a proposal set in $\mathcal{B}$. Specifically, we calculate the similarity score $a_{ij}$ between semantic role $s_i$ and proposal set $\mathcal{B}_j$ by: 
\begin{equation} \label{eq:4}
\begin{aligned}
\bm{q}_i = \left[ \bm{e}^g_v; \bm{e}^g_{s_i}; \bm{\bar{f}} \right], \quad  a_{ij} = F_a (\bm{q}_i, \bm{\bar{f}_j}),
\end{aligned}
\end{equation}
where $\bm{e}^g_v$ and $\bm{e}^g_{s_i}$ are the word embedding features of verb $v$ and semantic role $s_i$, $\bm{\bar{f}}$ and $\bm{\bar{f}_j}$ represent the average-pooled visual features of proposal set $\mathcal{B}$ and $\mathcal{B}_j$, [;] is a concatenation operation, and $F_a$ is a learnable similarity function\footnote{For conciseness, we leave the details in the supplementary material. \label{ftnote:concise}}.

After obtaining the grounding similarity scores $\{a_{ij}\}$ between semantic role $s_i$ and all proposal sets $\{\mathcal{B}_j\}$, we then select the top $n_i$ proposal sets with the highest scores as the grounding results for all sub-roles of $s_i$. $\mathcal{\tilde{R}}$ in Eq.~\eqref{eq:3} is the set of visual features of all grounded proposal sets.

\subsubsection{Semantic Structure Planner (SSP)}

Semantic structure planner (SSP) is a hierarchical semantic structure learning model, which aims to learn a reasonable sequence of sub-roles $\mathcal{S}$. As shown in Figure~\ref{fig:architecture}, it consists of two subnets: an S-level SSP and an R-level SSP.

\noindent\textbf{S-level SSP.} The sentence-level (S-level) SSP is a coarse-grained structure learning model, which only learns a sequence of all involved general semantic roles (including the verb) in $\mathcal{VSR}$ (\eg, \texttt{ride}, \texttt{Arg0}$_\text{rider}$, \texttt{Arg1}$_\text{steed}$ and \texttt{LOC}  in Figure~\ref{fig:architecture}). To this end, we formulate this sentence-level structure learning as a role sequence generation task, as long as we constrain that each output role token belongs to the given role set and each role can only appear once. Specifically, we utilize a three-layer Transformer~\cite{vaswani2017attention}\footnote{More comparison results between Transformer and Sinkhorn networks~\cite{mena2018learning,cornia2019show} are left in supplementary material.} to calucate the probability of roles $p(s_t | \mathcal{VSR})$ at each time step $t$\footref{ftnote:concise}: 
\begin{equation}
\begin{aligned}
\bm{H} & = \text{Transformer}_{\text{enc}} \left(\{\text{FC}_a(\bm{e}^i_v + \bm{e}^i_{s_i})\} \right), \\
p(s_t | \mathcal{VSR}) & = \text{Transformer}_{\text{dec}} \left( \bm{H}, \bm{e}^o_{s_{<t}} \right), \\
\end{aligned}
\end{equation}
where Transformer$_*$ are the encoder (enc) and decoder (dec) of the standard multi-head transformer. $\bm{e}^i_v$ and $\bm{e}^i_{s_i}$ are the word embedding features of verb $v$ and semantic role $s_j$, respectively. $\text{FC}_a$ is a learnable fc-layer to obtain the embedding of each input token. $\bm{e}^o_{s_{<t}}$ is the sequence of embeddings of previous roles. Based on $p(s_t | \mathcal{VSR})$, we can predict a role at time step $t$ and obtain an initial role sequence, \eg, \texttt{Arg0}$_\text{rider}$ -- \texttt{ride} -- \texttt{Arg1}$_\text{steed}$ -- \texttt{LOC} in Figure~\ref{fig:architecture}.

\noindent\textbf{R-level SSP.} The role-level (R-level) SSP is a fine-grained structure model which aims to rank all sub-roles within the same semantic role (\eg, \textcolor[rgb]{0.74, 0.2, 0.64}{\texttt{LOC-1}} and \textcolor[rgb]{0.13, 0.55, 0.13}{\texttt{LOC-2}} are two sub-roles of role \texttt{Loc} in Figure~\ref{fig:architecture}). Since the only differences among these sub-roles are the grounded visual regions, we borrow ideas from the Sinkhorn networks~\cite{mena2018learning,cornia2019show}, which use a differentiable Sinkhorn operation to learn a \emph{soft} permutation matrix $\bm{P}$. Specifically, for each role $s_i$ with multiple sub-roles (\ie, $n_i>1$), we first select all the corresponding grounded proposal sets for these sub-roles, denoted as $\mathcal{\hat{B}} = \{\mathcal{\hat{B}}_1, ..., \mathcal{\hat{B}}_{n_i}\}$. And for each proposal $\bm{b}_* \in \mathcal{\hat{B}}$, we encode a feature vector $\bm{z}_* = [\bm{z}^v_*; \bm{z}^{s_i}_*; \bm{z}^l_*]$, where $\bm{z}^v_*$ is a transformation of its visual feature $\bm{f}_*$, $\bm{z}^{s_i}_*$ is the word embedding feature of the semantic role $s_i$, and $\bm{z}^l_*$ is a 4-d encoding of the spatial position of proposal $\bm{b}_*$. Then, we transform each feature $\bm{z}_*$ into $n_i$-d, and average-pooled all features among the same proposal set, \ie, we can obtain an $n_i$-d feature for each $\mathcal{\hat{B}}_i$. We concatenate all these features to get an $n_i \times n_i$ matrix $\bm{Z}$. Finally, we use the Sinkhorn operation to obtain the soft permutation matrix $\bm{P}$\footref{ftnote:concise}:
\begin{equation}
\begin{aligned}
\bm{P} = \text{Sinkhorn}(\bm{Z}).
\end{aligned}
\end{equation}

After the two SSP subnets (\ie, S-level and R-level), we can obtain the semantic structure $\mathcal{S}$ (cf. Eq.~\eqref{eq:3}). Based on the sequence of $\mathcal{S}$ and the set of proposal featurs $\mathcal{\tilde{R}}$ from the GSRL model, we re-rank $\mathcal{\tilde{R}}$ based on  $\mathcal{S}$ and obtain $\mathcal{R}$.

\subsubsection{Role-shift Caption Generation}

Given the semantic structure sequence $\mathcal{S} = (s^b_1, ..., s^b_K)$ and corresponding proposal feature sequence $\mathcal{R} = (\bm{r}_1, ..., \bm{r}_K)$, we utilize a two-layer LSTM to generate the final caption $\bm{y}$. At each time step, the model fouces on one specific sub-role $\bm{s}^b_t$ and its grounded region set $\bm{r}_t$, and then generates the word $y_t$. Therefore, we take inspirations from previous CIC methods~\cite{cornia2019show,chen2020say}, and predict two distributions simultaneously: $p(g_t| \mathcal{S}, \mathcal{R})$ for controlling the shift of sub-roles, and $p(y_t|\mathcal{S}, \mathcal{R})$ to predict the distribution of a word.

As for the role-shift, we use an adaptive attention mechanism~\cite{lu2017knowing} to predict the probability of shifting\footref{ftnote:concise}:
\begin{equation}
\begin{aligned}
\alpha^g_t, \bm{\alpha}^r_t, \bm{sr}^g_t & = \text{AdaptiveAttn}_a (\bm{x}_t, \bm{r}_t),
\end{aligned}
\end{equation}
where $\text{AdaptiveAttn}_a$ is an adaptive attention network, $\bm{x}_t$ is the input query for attention, $\bm{sr}^g_t$ is a sential vector, $\alpha^g_t$ and $\bm{a}^r_t$ are the attention weights for the sential vector and region features, respectively. We directly use attention weight $\alpha^g_t$ as the probability of shifting sub-roles, \ie, $p(g_t | \mathcal{S},\mathcal{R}) = \alpha^g_t$. Based on probability $p(g_t| \mathcal{S}, \mathcal{R})$, we can sample a gate value $g_j \in \{0, 1\}$, and the focused sub-role at time step $t$ is:
\begin{equation}
\begin{aligned}
s^b_t \leftarrow \mathcal{S}[i], \text{where} \;  i = \min \left(1 + \textstyle{\sum}^{t-1}_{j=1} g_j, K \right).
\end{aligned}
\end{equation}
Due to the special nature of sub-role ``verb", we fix $g_{t+1}=1$ when $s^b_t$ is the verb.

For each sub-role $s^b_t$, we use the corresponding proposal set features $\bm{r}_t$ and a two-layer LSTM to generate word $y_t$:
\begin{equation}
\begin{aligned}
\bm{h}^1_t & = \text{LSTM}_1 \left(\bm{h}^1_{t-1}, \{y_{t-1}, \bm{\bar{f}}, \bm{h}^2_{t-1}\} \right), \\
\bm{h}^2_t & = \text{LSTM}_2 \left(\bm{h}^2_{t-1}, \{\bm{h}^1_t, \bm{c}_t \}\right), \\
y_t & \sim p(y_t| \mathcal{S}, \mathcal{R}) = \text{FC}_b (\bm{h}^2_t),
\end{aligned}
\end{equation}
where $\bm{h}^1_t$ and $\bm{h}^2_t$ are hidden states of the first- and second-layer LSTM (\ie, \text{LSTM}$_1$ and \text{LSTM}$_2$), $\text{FC}_b$ is a learnable fc-layer, and $\bm{c}_t$ is a context vector. To further distinguish the textual and visual words, we use another adaptive attention network to obtain the context vector $\bm{c}_t$\footref{ftnote:concise}:
\begin{equation}
\begin{aligned}
\alpha^v_t, \bm{\alpha}^r_t, \bm{sr}^v_t & = \text{AdaptiveAttn}_b (\bm{x}_t, \bm{r}_t), \\
\bm{c}_t & = \alpha^v_t  \cdot \bm{sr}^v_t  + \textstyle{\sum}_i \bm{\alpha}^r_{t, i} \cdot \bm{r}_{t, i}, \\
\end{aligned}
\end{equation}
where $\bm{x}_t$ is the query for adaptive attention (\ie, the input of the $\text{LSTM}_1$), $\bm{sr}^v_t$ is a sential vector, and $\alpha^v_t$ and $\bm{\alpha}^r_t$ are the attention weights for the sential vector and region features.

\subsection{Training and Inference} \label{sec:traintest}

\noindent\textbf{Training Stage.} In the training stage, we train the three components (GSRL, SSP and captioning model) separately:

\noindent\emph{Training objective of GSRL.} For the GSRL model, we use a binary cross-entropy (BCE) loss between the predicted similarity scores $\hat{a}_{ij}$ and its ground truth $a^*_{ij}$ as the training loss:
\begin{equation}
\begin{aligned}
L_{\text{GSRL}} = \textstyle{\sum}_{ij} \text{BCE}(\hat{a}_{ij}, a^*_{ij}).
\end{aligned}
\end{equation}

\noindent\emph{Training objective of SSP.} For S-level SSP, we use a cross-entropy (XE) loss between prediction $\hat{s}_t$ and its ground truth $s^*_t$ as the training objective. For R-level SSP, we use a mean square (MSE) loss between prediction $\bm{\hat{P}}_t$ and its ground truth $\bm{P^*}_t$ as the training objective:
\begin{equation}
\begin{aligned}
L^S_{\text{SSP}} = \textstyle{\sum}_t \text{XE}(\hat{s}_t, s^*_t), L^R_{\text{SSP}} = \textstyle{\sum}_t \mathbf{1}_{(n_t > 1)} \text{MSE}(\bm{\hat{P}}_t, \bm{P^*}_t),
\end{aligned}
\end{equation}
where $\mathbf{1}_{(n_t > 1)}$ is an indicator function, being 1 if $n_t > 1$ and 0 otherwise.

\noindent\emph{Training objective of captioning model.} We follow the conventions of previous captioning works and use a two-stage training scheme: XE and RL stages. In the XE stage, we use an XE loss between predicted words and ground truth words as the training loss. In the RL stage, we use a self-critical baseline~\cite{rennie2017self}. At each step, we sample from $p(y_t | \mathcal{S}, \mathcal{R})$ and $p(g_t|\mathcal{S}, \mathcal{R})$ to obtain the next word $y_{t+1}$ and sub-role $s^b_{t+1}$. Then we calcuate the reward $r(\bm{y}^s)$ of the sampled sentence $\bm{y}^s$. Baseline $b$ is the reward of the greedily generated sentence. Thus, the gradient expression of the training loss is:
\begin{equation}
\nabla_{\theta} L = - (r(\bm{y}^s) - b) ( \nabla_{\theta} \log p(\bm{y}^s) + \nabla_{\theta} \log p(\bm{g}^s) ),
\end{equation}
where $\bm{g}^s$ is the sequence of role-shift gates.

\begin{figure}[t]
	\centering
	\includegraphics[width=0.82\linewidth]{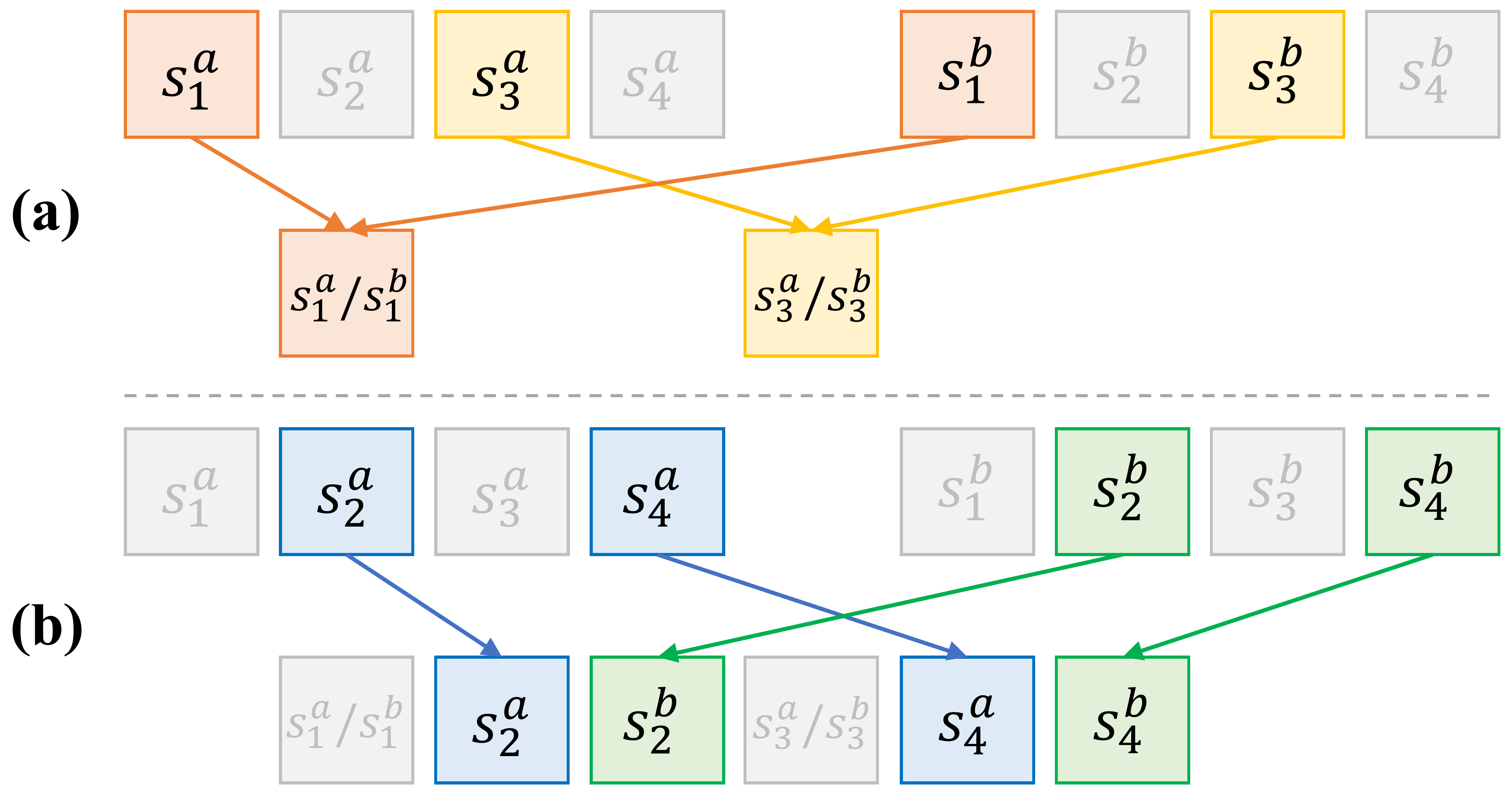}
	\vspace{-1em}
	\caption{A toy example of merging two different semantic structures $\mathcal{S}^a$ and $\mathcal{S}^b$ into a single sub-role sequence.}
	\vspace{-1em}
	\label{fig:merge}
\end{figure}

\noindent\textbf{Inference.} In testing stage, given an image and one $\mathcal{VSR}$, we sequentially use the GSRL, SSP, and captioning model to generate the final captions. Meanwhile, our framework can be easily extended from one $\mathcal{VSR}$ to multiple $\mathcal{VSR}s$ as the control signal. Taking an example of two $\mathcal{VSR}s$, we first use GSRL and SSP to obtain semantic structures and grounded regions features: $(\mathcal{S}^a, \mathcal{R}^a)$ and $(\mathcal{S}^b, \mathcal{R}^b)$. Then, as shown in Figure~\ref{fig:merge}, we merge them by two steps\footref{ftnote:concise}: (a) find the sub-roles in both $\mathcal{S}^a$ and $\mathcal{S}^b$ which refer to the same visual regions (\eg, $s^a_1$ and $s^b_1$ refer to the same proposal set); (b) insert all other sub-roles between the nearest two selected sub-roles (\eg, $s^*_2$ are still between $s^*_1$ and $s^*_3$). Concerning the order of sub-roles from different verbs, we follow the rank of two verbs (\eg, $s^a_2$ is in front of $s^b_2$).

\section{Experiments}

\subsection{Datasets and Metrics}

\noindent\textbf{Flickr30K Entities~\cite{plummer2015flickr30k}.} It builds upon the Flickr30K~\cite{young2014image} dataset, by manually grounding each noun phrase in the descriptions with one or more visual regions. It consists of 31,000 images, and each image is associated with five captions. We use the same splits as~\cite{karpathy2015deep} in our experiments.

\noindent\textbf{COCO Entities~\cite{cornia2019show}.} It builds upon the COCO~\cite{chen2015microsoft} dataset which consists of 120,000 images and each image is annotated with five captions. Different from Flickr30K Entities where all grounding entities are annotated by humans, all annotations in COCO Entities are detected automatically. Especially, they align each entity to all the detected proposals with the same object class.

Although we only assume that there exists at least one verb (\ie, activity) in each image; unfortunately, there are still a few samples (\ie, 3.26\% in COCO Entities and 0.04\% in Flickr30K Entities) having no verbs in their captions. We use the same split as~\cite{cornia2019show} and further drop the those samples with no verb in the training and testing stages\footref{ftnote:concise}. We will try to cover these extreme cases and leave it for future work.

\subsection{Implementation Details} \label{sec:details}

\noindent\textbf{Proposal Generation and Grouping.} We utilize a Faster R-CNN~\cite{ren2016faster} with ResNet-101~\cite{he2016deep} to obtain all proposals for each image. Especially, we use the model released by~\cite{anderson2018bottom}, which is finetuned on VG dataset~\cite{krishna2017visual}. For COCO Entities, since the ``ground truth" annotations for each noun phrase are the proposals with the same class, we group the proposals by their detected class labels. But for Flickr30K Entities, we directly regard each proposal as a proposal set.

\noindent\textbf{VSR Annotations.} Since there are no ground truth semantic role annotations for CIC datasets, we use a pretrained SRL tool~\cite{shi2019simple} to annotate verbs and semantic roles for each caption, and regard them as ground truth annotations. For each detected verb, we convert it into its base form and build a verb dictionary for each dataset. The dictionary sizes for COCO and Flickr30K are 2,662 and 2,926, respectively. There are a total of 24 types of semantic roles for all verbs.

\noindent\textbf{Experimental Settings.}
For the S-level SSP, the head number of multi-head attention is set to 8, and the hidden size of the transformer is set to 512. The length of the transformer is set to 10.
For the R-level SSP, we set the maximum number of entities for each role to 10.
For the RL training of the captioning model, we use CIDEr-D~\cite{vedantam2015cider} score as the training reward. Due to the limited space, we leave more detailed parameter settings in the supplementary material.

\addtolength{\tabcolsep}{-2pt}
\begin{table*}
	\small
	\begin{center}
		\scalebox{0.92}{
			\begin{tabular}{| l | c | c c c c c| c c c | c c c c c| c c c |}
				\hline
				\multirow{2}{*}{Method}  & 	\multirow{2}{*}{Proposals} & \multicolumn{8}{c|}{COCO Entities}  & \multicolumn{8}{c|}{Flickr30K Entities} \\
				\cline{3-18}
				& & B4 & M & R & C & S & R$_\text{V}$ & R$_\text{SR1}$ & R$_\text{SR2}$ &  B4 & M & R & C & S & R$_\text{V}$ & R$_\text{SR1}$ & R$_\text{SR2}$ \\
				\hline
				C-LSTM~\cite{vinyals2015show} & GSRL & 12.5 & 19.1 & 41.4 & 126.5 & 31.5 & 29.6 & 20.7 & 14.3 & 7.4 & 13.1 & 31.0 & 58.7 & 18.9 & 16.9 & 14.5 & 9.1 \\
				C-UpDn~\cite{anderson2018bottom} & GSRL & 13.4 & 19.9 & 42.3 & 135.7 & 32.9 & 30.4 & 21.1 & 13.5 & 7.5 & 13.0 & 31.4 & 58.7 & \textbf{18.9} & 16.2 & 13.9 & 9.0 \\
				SCT~\cite{cornia2019show} & GSRL & 12.4 & 19.0 & 42.1 & 127.6 & 34.6 & 28.7 & 19.5 & 14.8 & 6.9 & 12.7 & 29.7 & 50.6 & 17.7 & 16.1 & 13.7 & 9.1 \\
				Ours~\textit{w/o} verb & GSRL & 13.4 & 19.2 & 42.8 & 129.5 & 34.7 & 30.4 & 21.0 & 15.9 & 7.0 & 12.7 & 29.7 & 50.8 & 17.6 & 16.8 & 14.5 & 9.5 \\
				\textbf{Ours} & GSRL & \textbf{16.0} & \textbf{23.2} & \textbf{47.1} & \textbf{162.8} & \textbf{35.7} & \textbf{81.3} & \textbf{54.3} & \textbf{36.3} & \textbf{7.9} & \textbf{14.7} & \textbf{32.6} & \textbf{71.6} & 18.2 & \textbf{49.8} & \textbf{38.0} & \textbf{24.7} \\
				\textcolor{mygray}{Ours (oracle verb)} & \textcolor{mygray}{GSRL} & \textcolor{mygray}{17.5} & \textcolor{mygray}{24.0} & \textcolor{mygray}{49.0} & \textcolor{mygray}{184.3} & \textcolor{mygray}{35.7} & \textcolor{mygray}{96.8} & \textcolor{mygray}{64.5} & \textcolor{mygray}{43.6} & \textcolor{mygray}{9.0} & \textcolor{mygray}{16.0} & \textcolor{mygray}{35.4} & \textcolor{mygray}{96.5} & \textcolor{mygray}{18.6} & \textcolor{mygray}{73.3} & \textcolor{mygray}{55.5} & \textcolor{mygray}{36.1}  \\
				\hline
				C-LSTM~\cite{vinyals2015show} & GT & 14.6 & 21.1 & 44.3 & 148.2 & 36.3 & 29.7 & 20.7 & 14.2 & 9.3 & 14.7 & 34.3 & 75.7 & 22.4 & 17.0 & 14.8 & 9.6 \\
				C-UpDn~\cite{anderson2018bottom} & GT & 16.5 & 22.9 & 46.7 & 170.0 & 40.4 & 30.5 & 21.3 & 13.6  & 9.4 & 14.7 & 34.5 & 74.8 & 22.5 & 16.2 & 14.0 & 9.2   \\
				SCT~\cite{cornia2019show} & GT & 18.1 & 24.4 & 50.3 & 191.3 & 47.4 & 29.4 & 20.1 & 15.3 & 10.1 & 15.9 & 36.3 & 82.0 & \textbf{24.3} & 17.2 & 14.8 & 9.7 \\
				Ours~\textit{w/o} verb & GT &  20.1 & 24.3 & 52.8 & 199.5 & 47.3 & 30.6 & 21.0 & 16.2 & 10.3 & 15.8 & 36.8 & 82.2 & 24.0 & 17.8 & 15.3 & 10.4  \\
				\textbf{Ours} & GT & \textbf{23.1} & \textbf{28.0} & \textbf{55.6} & \textbf{235.1} & \textbf{48.9} & \textbf{71.2} & \textbf{47.8} & \textbf{34.2} & \textbf{10.7} & \textbf{18.0} & \textbf{37.1} & \textbf{97.5} & 21.9 & \textbf{57.9} & \textbf{44.7} & \textbf{28.6} \\
				\textcolor{mygray}{Ours (oracle verb)} & \textcolor{mygray}{GT} & \textcolor{mygray}{25.4} & \textcolor{mygray}{28.8} & \textcolor{mygray}{57.8} & \textcolor{mygray}{265.0} & \textcolor{mygray}{49.8} & \textcolor{mygray}{88.0} & \textcolor{mygray}{59.2} & \textcolor{mygray}{42.5} & \textcolor{mygray}{12.3} & \textcolor{mygray}{19.8} & \textcolor{mygray}{40.9} & \textcolor{mygray}{131.4} & \textcolor{mygray}{22.4}  & \textcolor{mygray}{86.2} & \textcolor{mygray}{66.2} & \textcolor{mygray}{42.5} \\
				\hline
				
			\end{tabular}
		} 
	\end{center}
	\vspace{-2.0em}
	\caption[]{Performance (\%) compared with SOTA methods for controllable image captioning. The upper part denotes that all grounded proposal sets come from the GSRL model, and the below part denotes that all grounded proposal sets come from ground truth annotations.}
	\vspace{-1.0em}
	\label{tab:SOTA}
\end{table*}
\addtolength{\tabcolsep}{2pt}

\begin{figure*}[h]
	\centering
	\begin{minipage}[h]{0.43\linewidth}
		\includegraphics[width=0.95\linewidth]{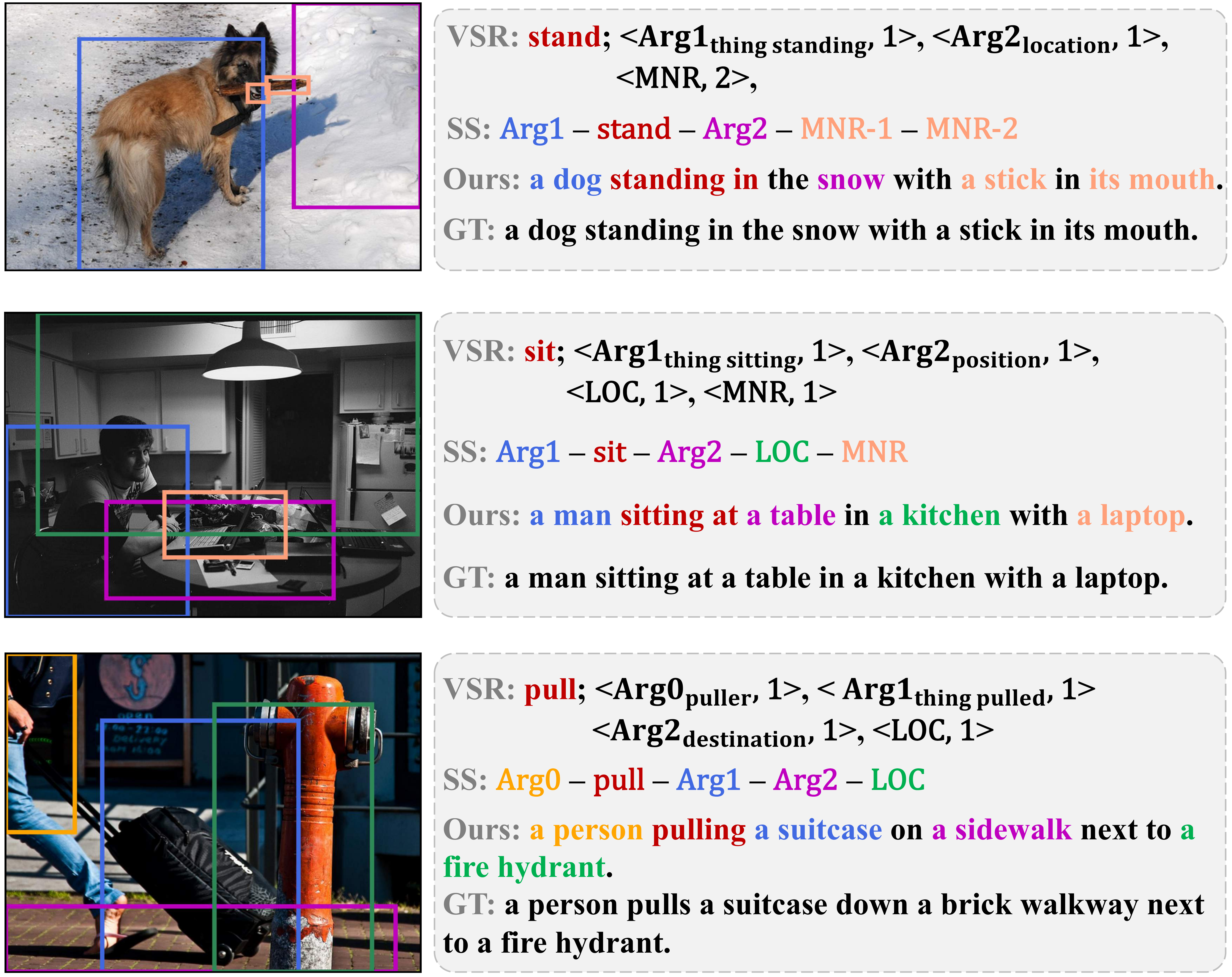}
		\vspace{-1.0em}
		\caption{Examples of generated image captions using the VSR corresponding to the ground truth caption. \textbf{SS} denotes the learned semantic structures.}
		\vspace{-0.5em}
		\label{fig:vis-a}
	\end{minipage}
	\quad
	\begin{minipage}[h]{0.54\linewidth}
		\includegraphics[width=0.95\linewidth]{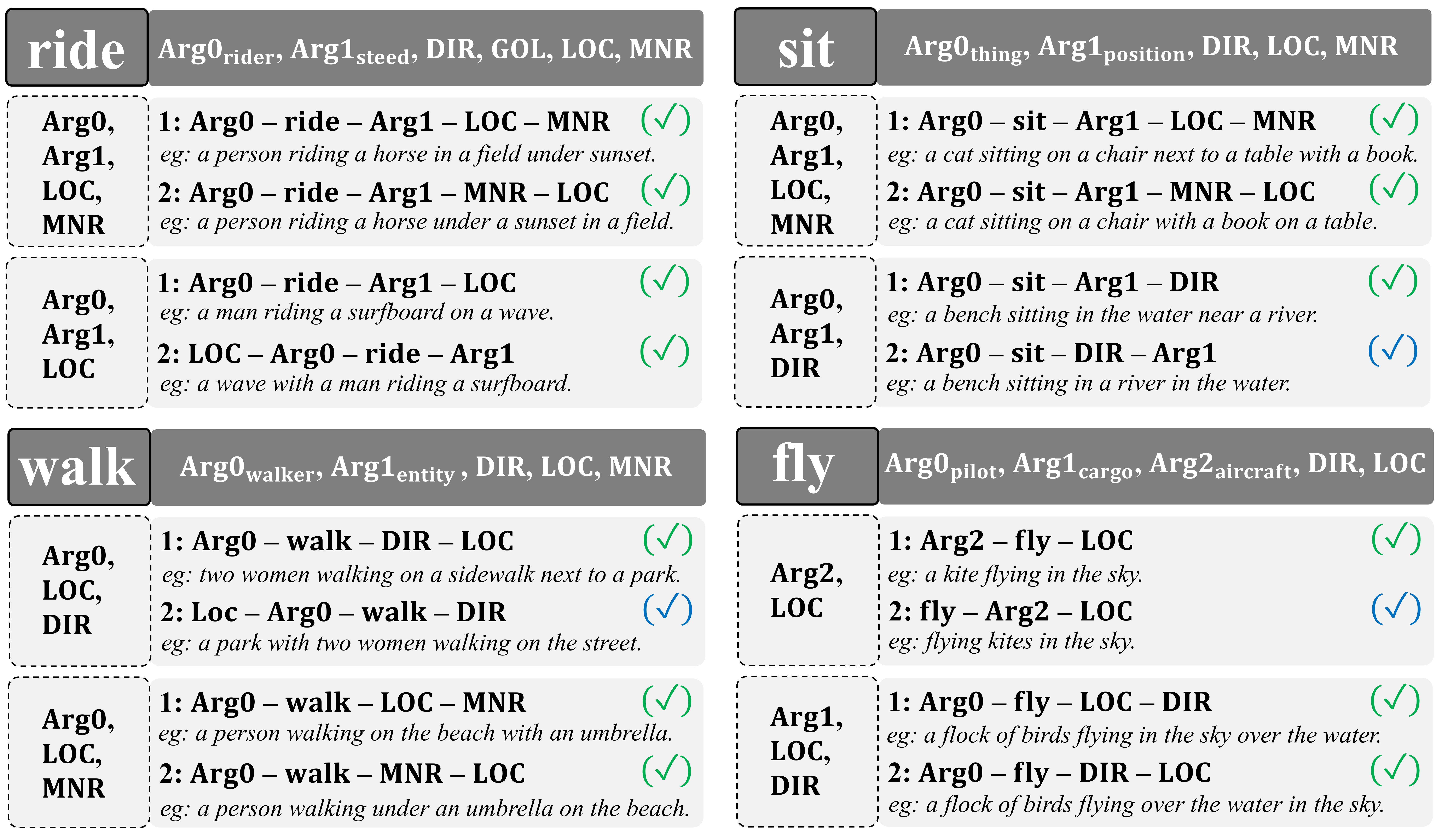}
		\vspace{-1.0em}
		\caption{Examples of the learned verb-specific semantic structures. The first row of each sample is the verb and all reasonable semantic roles. The second or third row is a sampled role set with two top-ranking structures. The green (blue) tick denotes that this structure is (not) in the dataset.}
		\vspace{-0.5em}
		\label{fig:vis-b}
	\end{minipage}
\end{figure*}

\subsection{Evaluation on Controllability}

\noindent\textbf{Settings.} To evaluate the controllability of proposed framework, we followed the conventions of prior CIC works~\cite{cornia2019show,chen2020say,zhong2020comprehensive}, and utilized the VSR aligned with ground truth captions as the control signals. Specifically, we compared the proposed framework with several carefully designed baselines\footnote{All baselines use the same visual regions as models with VSRs.}: 1) \textbf{C-LSTM}: It is a Controllable LSTM model~\cite{vinyals2015show}. Given the features of all grounded visual regions, it first averages all region features, and then uses an LSTM to generate the captions. 2) \textbf{C-UpDn}: It is a Controllable UpDn model~\cite{anderson2018bottom}, which uses an adaptive attention to generate the captions. 3) \textbf{SCT}~\cite{cornia2019show}: It regards the set of visual regions as a control signal, and utilizes a chunk-shift captioning model to generate the captions. 4) \textbf{Ours \emph{w/o} verb}: We ablate our model by removing the verb information in both the SSP and captioning model. 5) \textbf{Ours (oracle verb)}: It is an ideal situation, where the captioning model directly outputs the oracle format of the verb when the attending role is the verb.

\noindent\textbf{Evaluation Metrics.} To evaluate the quality of the generated captions, we use five accuracy-based metrics, including BLEU-4 (B4)~\cite{papineni2002bleu}, METEOR (M)~\cite{banerjee2005meteor}, ROUGE (R)~\cite{lin2004rouge}, CIDEr-D (C)~\cite{vedantam2015cider}, and SPICE (S)~\cite{anderson2016spice}. Particularly, we evaluate the generated captions against the single ground truth caption. We also propose a new recall-based metric to evaluate whether the roles of the generated sentence are consistent with the ground truth caption (\ie, VSR). It measures the recall rate of the verb, semantic roles, and ordered role pairs, which are denoted as R$_{\text{V}}$, R$_{\text{SR1}}$ and R$_{\text{SR2}}$, respectively.

\begin{figure*}[t]
	\centering
	\includegraphics[width=0.95\linewidth]{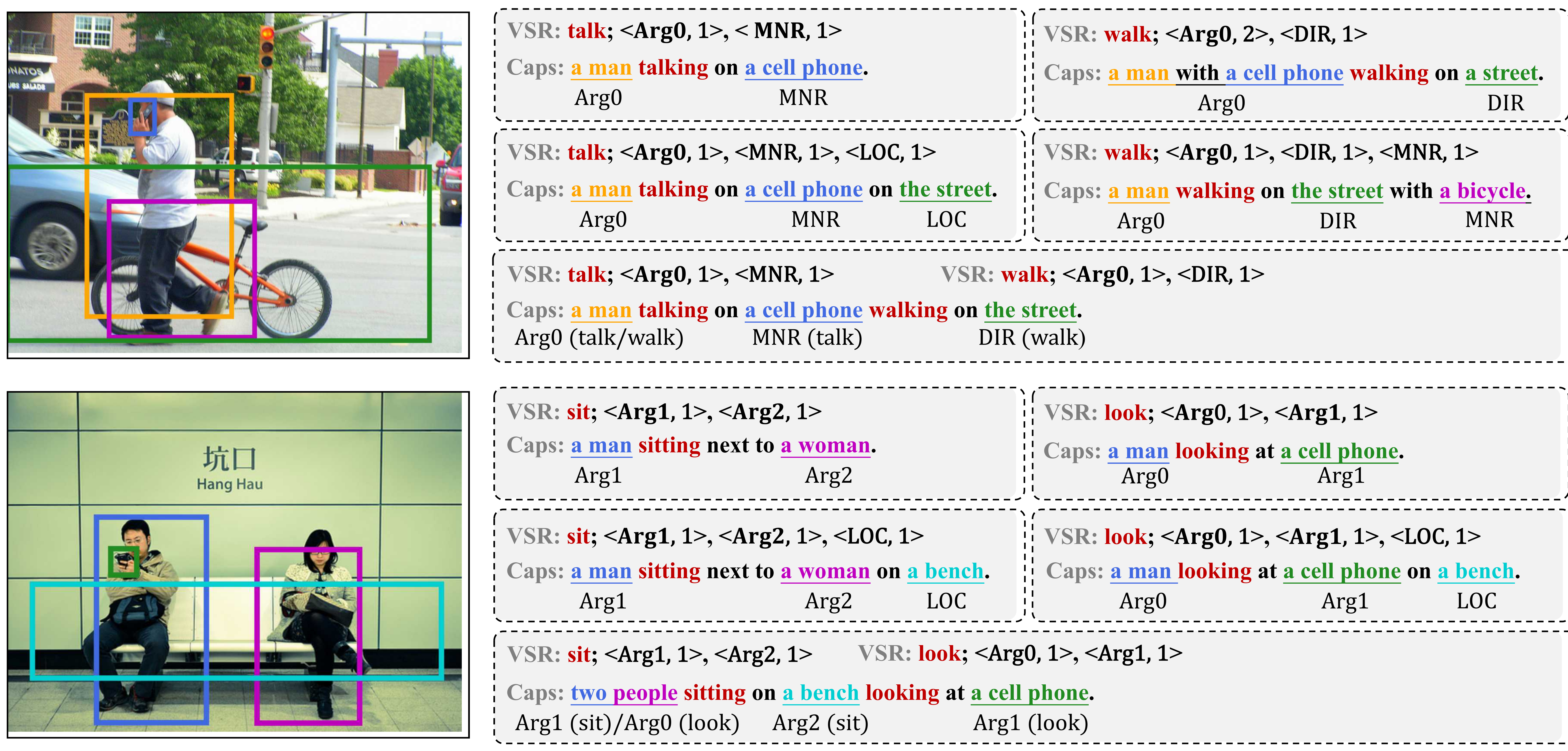}
	\vspace{-1.0em}
	\caption{Examples of diverse image caption generation conditioned on different VSRs. Best viewed in color.}
	\vspace{-1.2em}
	\label{fig:vis-c}
\end{figure*}

\noindent\textbf{Quantitative Results.} The quantitative results are reported in Table~\ref{tab:SOTA}. From Table~\ref{tab:SOTA}, we can observe that our framework can achieve the best performance over almost all metrics and benchmarks. By comparing the two different proposal settings (\ie, GSRL and GT), we can find that the accuracy of GSRL is a major bottleneck of the whole framework. Meanwhile, the ablative model (Ours \emph{w/o} verb) can only achieve slightly better performance than baseline SCT and much worse performance than our full model, which reflects the importance of the verb in semantic structure learning and caption generation.

\noindent\textbf{Visualizations.} In Figure~\ref{fig:vis-a}, we illustrate some examples of the generated captions. We can observe that our framework always learns a human-like semantic structure based on the VSR and grounded visual regions (\eg, \texttt{Arg1}$_\text{thing}$ -- \texttt{sit} -- \texttt{Arg2}$_\text{position}$ -- \texttt{LOC} -- \texttt{MNR}). According to the semantic structures, the captioning model can generate near-perfect descriptions. As a by-product, a well-trained SSP can automatically produce several verb-specific semantic structures for a set of user-interested roles, and we show some examples in Figure~\ref{fig:vis-b}. For each verb and role set, we illustrate the top two structures by using beam search. Particularly, we are surprised to find that we can even learn some structures that never appear in original datasets (the blue tick ones).

\subsection{Evaluation on Diversity}
One of the well-known advantages of controllable image captioning is the ability to generate diverse image captions by feeding different control signals. Thus, we also evaluate the diversity of the captions generated by our framework.

\noindent\textbf{Settings.} We evaluated the quality of diverse captions in two settings: 1) Given a VSR and grounded visual regions of each role aligned with the ground truth caption, we first use an SSP to select two semantic structures, and then respectively generate two diverse captions. For fair comparisons, we utilize the same set of visual regions on two strong baselines: a) \textbf{BS}: an UpDn model uses beam search to produce two captions, and b) \textbf{SCT}: an SCT model takes a permutation of all region sets to generate two captions. 2) For each verb, we can randomly sample a subset of all semantic roles to construct new VSRs. Specifically, we sample two more sets of semantic roles, and generate two diverse captions for each role set following the same manner. 

\addtolength{\tabcolsep}{-2pt}
\begin{table}[t]
	\small
	\begin{center}
		\scalebox{0.95}{
			\begin{tabular}{| l | c | c c c c c| c c c |}
				\hline
				\multirow{2}{*}{Model} & \multirow{2}{*}{\#caps} & \multicolumn{5}{c|}{Accuracy-based} & \multicolumn{3}{c|}{Diversity-based} \\
				&  & B4 & M & R & C & S & D-1 & D-2 & s-C \\
				\hline
				BS & 2 & 18.1 & 24.0 & 48.8 & 185.7 & 43.7 & 45.5 & 61.2 & 53.5  \\
				SCT & 2 & 20.5 & 25.8 & 53.0 & 210.0 & 51.6 & \textbf{52.2} & \textbf{73.7} & \textbf{76.0}  \\
				Ours & 2 & \textbf{24.8} & \textbf{29.5} & \textbf{57.8} & \textbf{251.9} & \textbf{53.1} & 48.3 & 70.0 & 68.3  \\
				\hline
				BS & 6 & 20.9 & 25.4 & 52.1 & 209.5 & 47.9 & 22.7 & 35.6 & 53.9  \\
				SCT & 6  & 22.0 & 26.5 & 55.4 & 222.5 & 54.9 & \textbf{27.7} & \textbf{45.7} & \textbf{69.1}  \\
				Ours & 6  & \textbf{26.6} & \textbf{30.2} & \textbf{59.8} & \textbf{267.3} & \textbf{56.6} & 25.1 & 43.8 & 67.0  \\
				\hline
			\end{tabular}
		} 
	\end{center}
	\vspace{-2.0em}
	\caption[]{Performance compared with two strong baselines for diverse image captioning on dataset COCO Entities.}
	\vspace{-1em}
	\label{tab:diversity}
\end{table}
\addtolength{\tabcolsep}{2pt}

\noindent\textbf{Evaluation Metrics.} We used two types of metrics to evaluate the diverse captions: 1) Accuracy-based: we followed the conventions of the previous works~\cite{cornia2019show,deshpande2019fast,wang2017diverse} and reported the best-1 accuracy, \ie, the generated caption with the maximum score for each metric is chosen. Analogously, we evaluate the generated captions against the single ground truth caption. 2) Diversity-based: we followed~\cite{chen2020say} and used two metrics which only focus on the language similarity: Div-n (D-n)~\cite{aneja2019sequential,deshpande2019fast} and self-CIDEr (s-C)~\cite{wang2019describing}.

\noindent\textbf{Quantitative Results.} The quantitative results are reported in Table~\ref{tab:diversity}. From Table~\ref{tab:diversity}, we can observe that the diverse captions generated by our framework in both two settings have much higher accuracy (\eg, CIDEr 267.3 vs. 222.5 in SCT), and that the diversity is slightly behind SCT (\eg, self-CIDEr 67.0 vs. 69.1 in SCT). This is because SCT generates captions by randomly shuffling regions. Instead, we tend to learn more reasonable structures. Thus, we can achieve much higher results on accuracy, \ie, our method can achieve a better trade-off between quality and diversity on diverse image captioning than the two strong baselines.

\noindent\textbf{Visualizations.} We further illustrate the generated captions of two images with different VSRs in Figure~\ref{fig:vis-c}. The captions are generated effectively according to the given VSR, and the diversity of VSR leads to significant diverse captions.

\section{Conclusions \& Future Works}

In this paper, we argued that all existing objective control signals for CIC have overlooked two indispensable characteristics: event-compatible and sample-suitable. To this end, we proposed a novel control signal called VSR. VSR consists of a verb and several semantic roles, \ie, all components are guaranteed to be event-compatible. Meanwhile, VSR only restricts the involved semantic roles, which is also sample-suitable for all the images containing the activity. We have validated the effectiveness of VSR through extensive experiments. Moving forward, we will plan to 1) design a more effective captioning model to benefit more from the VSR signals; 2) extend VSR to other controllable text generation tasks, \eg, video captioning~\cite{xu2018dual}; 3) design a more general framework to cover the images without verbs.

\footnotesize \noindent\textbf{Acknowledgements.}
This work was supported by the National Natural Science Foundation of China (U19B2043,61976185), Zhejiang Natural Science Foundation (LR19F020002), Zhejiang Innovation Foundation (2019R52002), and Fundamental Research Funds for Central Universities.

{\small
	\bibliographystyle{ieee_fullname}
	\bibliography{egbib}
}

\clearpage

\begin{normalsize}
	
\section*{Appendix}
\appendix

The supplementary document is organized as follows:
\begin{itemize}
	\item In Section~\ref{sec: supp1}, we explain the meanings of different semantic roles (\ie, PropBank-style annotations) in our paper.
	\item In Section~\ref{sec: supp2}, we illustrate more visualization results generated by our CIC framework.
	\item In Section~\ref{sec: supp3}, we provide the details about each subnet component of our VSR-guided CIC model.
	\item In Section~\ref{sec: supp4}, we show the details about the merging algorithm of two different semantic structures from two VSRs.
	\item In Section~\ref{sec: supp5}, we report the details of our experimental settings.
	\item In Section~\ref{sec: supp6}, we compare the performance between the Transformer structure and Sinkhorn network in S-level SSP.
\end{itemize}

\section{Meanings of Different Semantic Roles} \label{sec: supp1}

In this paper, we mainly follow the types of semantic roles defined in the PropBank~\cite{palmer2005proposition}. The main arguments with their semantic role meanings is listed in Table~\ref{tab: ARGS}, including numbered arguments (\eg, \texttt{Arg0}, \texttt{Arg2})\footnote{Since semantic role \texttt{Arg5} is very rare for the verbs of CIC datasets, and we omit it in Table~\ref{tab: ARGS}.} and argument modifiers (\eg, \texttt{COM}, \texttt{LOC}). 

\begin{table}
	\begin{center}
		\begin{tabular}{|c|c|c|} 
			\hline
			& Role Type & Meaning \\
			\hline
			\parbox[t]{2mm}{\multirow{5}{*}{\rotatebox[origin=c]{90}{numbered args}}} & \texttt{Arg0} & agent \\
			\cline{2-3}
			& \texttt{Arg1} & patient \\
			\cline{2-3}
			& \texttt{Arg2} & instrument, benefactive, attribute \\
			\cline{2-3}
			& \texttt{Arg3} & starting point, benefactive, attribute \\
			\cline{2-3}
			& \texttt{Arg4} & ending point \\
			\hline\hline
			\parbox[t]{2mm}{\multirow{18}{*}{\rotatebox[origin=c]{90}{argument modifiers}}} & \texttt{COM} & comitative \\
			\cline{2-3}
			& \texttt{LOC} & locative \\
			\cline{2-3}
			& \texttt{DIR} & directional \\
			\cline{2-3}
			& \texttt{GOL} & goal \\
			\cline{2-3}
			& \texttt{MNR} & manner \\
			\cline{2-3}
			& \texttt{TMP} & temporal \\
			\cline{2-3}
			& \texttt{EXT} & extent \\
			\cline{2-3}
			& \texttt{REC} & reciprocals \\
			\cline{2-3}
			& \texttt{PRD} & secondary predication \\
			\cline{2-3}
			& \texttt{PRP} & purpose \\
			\cline{2-3}
			& \texttt{PNC} & purpose not cause \\
			\cline{2-3}
			& \texttt{CAU} & cause \\
			\cline{2-3}
			& \texttt{DIS} & discourse \\
			\cline{2-3}
			& \texttt{ADV} & adverbials \\
			\cline{2-3}
			& \texttt{ADJ} & adjectival \\
			\cline{2-3}
			& \texttt{MOD} & modal \\
			\cline{2-3}
			& \texttt{NEG} & negation \\
			\cline{2-3}
			& \texttt{LVB} & light verb \\
			\hline
		\end{tabular}
	\end{center}
	\caption{List of the main arguments in the PropBank.}
	\label{tab: ARGS}
\end{table}

Although there are many kinds of arguments modifiers in the PropBank, the most common argument modifiers of the verbs in Flickr30k/COCO Entities are \texttt{LOC}, \texttt{DIR}, \texttt{GOL} and \texttt{MNR}. The meaning of them as listed as follows: 
\begin{itemize}
	\item \texttt{LOC}: Locative modifiers indicate where some action takes place. 
	\item \texttt{DIR}: Directional modifiers show motion along some path. 
	\item \texttt{GOL}: Goal tag is for the goal of the action of the verb. 
	\item \texttt{MNR}: Manner modifiers specify how an action is performed. 
\end{itemize}

\section{More Visualization Results} \label{sec: supp2}
We illustrate more visualization results of generated image captions using the VSR corresponding to the ground truth caption in Figure~\ref{fig:gt_example_2}. Meanwhile, we show more visualization results about diverse image captions conditioned on different VSRs in Figure~\ref{fig:visulazation_2}. More specifically, the VSRs in the top row of images contain the same verb and different semantic role sequences; the VSRs in the bottom row of images contain a different verb or two verbs.
%

\begin{figure*}[t]
	\centering
	\includegraphics[width=0.98\linewidth]{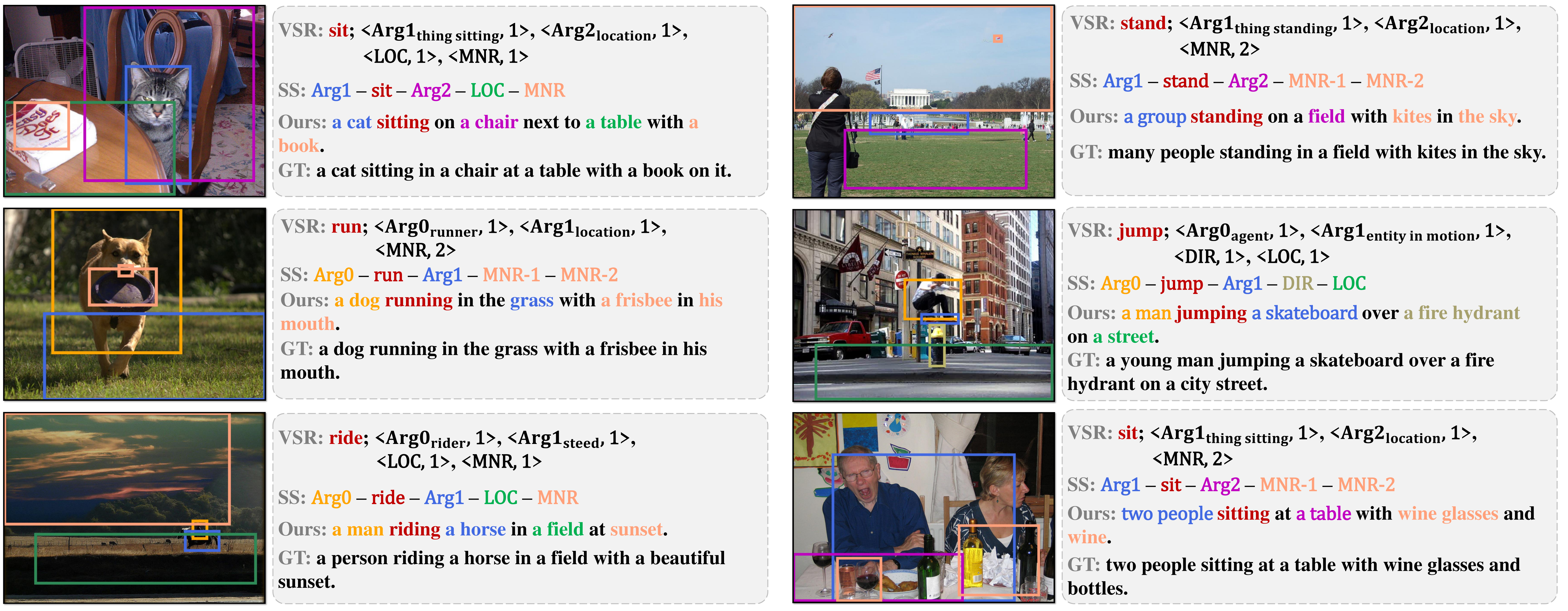}
	\caption{Additional examples of generated image captions using the VSR corresponding to the ground truth caption. \textbf{SS} denotes the learned semantic structures. Different colors show a correspondence between image regions and semantic roles. Best viewed in color.}
	\label{fig:gt_example_2}
\end{figure*}


\begin{figure*}[t]
	\centering
	\includegraphics[width=0.98\linewidth]{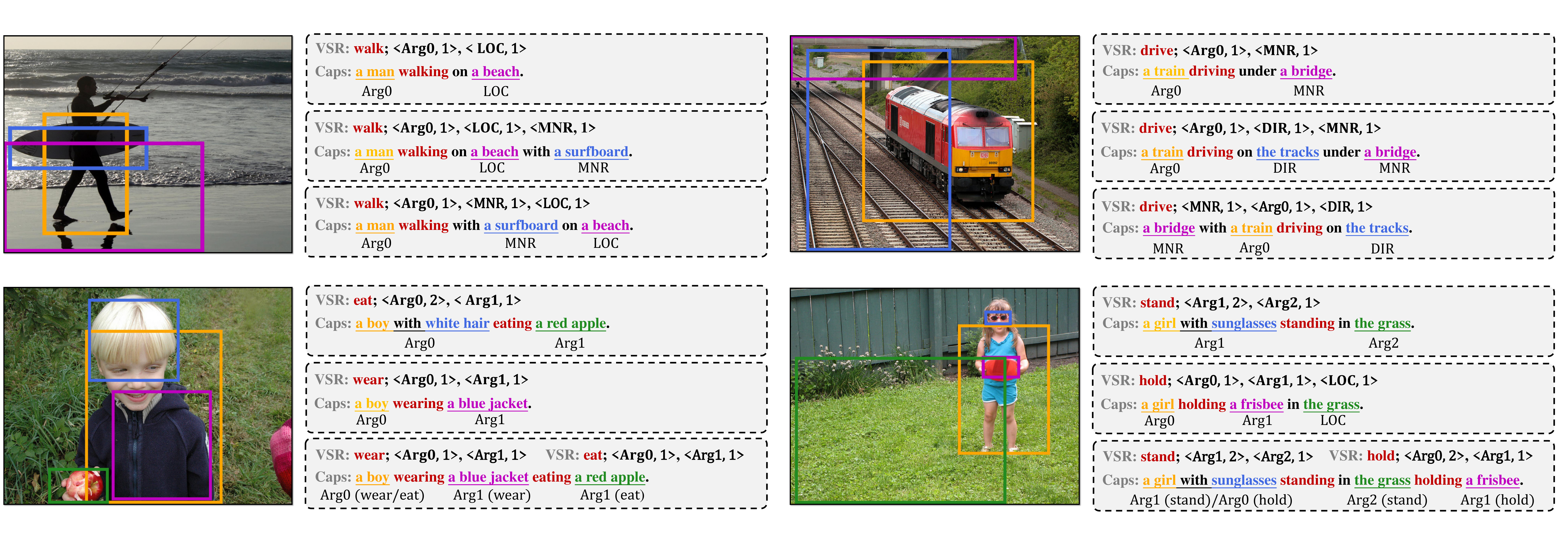}
	\caption{Additional examples of diverse image caption generation conditioned on different VSRs. The correspondences between image regions and noun phrases are indicated by different colors. Best viewed in color.}
	\label{fig:visulazation_2}
\end{figure*}


\section{Details of the VSR-guided CIC Model} \label{sec: supp3}

\subsection{Grounded Semantic Role Labeling}
In this grounded semantic role labeling (GSRL) step, we aim to ground each sub-role $s_i$ in $\mathcal{VSR}$ to a proposal set $\mathcal{B}_j \in \mathcal{B}$. Specifically, we calculate the similarity score $a_{ij}$ between sub-role $s_i$ and proposal set $\mathcal{B}_j$ by:
\begin{equation}
\begin{aligned}
\bm{q}_i & = [\bm{W}^g_v \Pi_v; \bm{W}^g_s \Pi_{s_i}; \bm{\bar{f}} ], \\
\bm{a}_{ij} & = \text{MLP}_a(\bm{W}^g_q \bm{q}_i \odot  \bm{W}^g_f \bm{\bar{f}_j}),
\end{aligned}
\end{equation}
where $\bm{\bar{f}} \in \mathbb{R}^{v \times 1}$ and $\bm{\bar{f}_j} \in \mathbb{R}^{v \times 1}$ represent the average-pooled visual feature of proposal set $\mathcal{B}$ and $\mathcal{B}_j$.  $\Pi_v$ and $\Pi_{s_i}$ are the one-hot embeddings for the verb $v$ and sub-role $s_i$, $\bm{W}^g_v \in \mathbb{R}^{d_v \times |\mathcal{V}|}$ and $\bm{W}^g_s \in \mathbb{R}^{d_s \times |\mathcal{SR}| }$ are learnable mapping matrices, $|\mathcal{V}|$ and $|\mathcal{SR}|$ are the size of the vocabulary of verbs and semantic roles, respectively. $[;]$ is a concatenation operation. Thus, $\bm{q}_i$ is a query vector combining the verb category, semantic role type and image global features. $\bm{W}^g_q \in \mathbb{R}^{a \times (d_v+d_s+v)}$ and $\bm{W}^g_f \in \mathbb{R}^{a \times v}$ aim to transform $\bm{q}_i$ and $\bm{\bar{f}_j}$ into a common space, and $\odot$ is the element-wise multiplication. Finally, a four-layer MLP maps the fused feature into a score $\bm{a}_{ij}$ between 0 and 1.

\subsection{Semantic Structure Planner}

\noindent\textbf{S-level SSP.} 
In the sentence-level (S-level) SSP, we utilize a three-layer Transformer encoder to encode the verb $v$ and semantic role $s_i$ in the input semantic role sequence $\mathcal{S}$. 
\begin{equation}
\bm{H} = \text{Transformer}_{\text{enc}} \left(\{\text{FC}_a(\bm{W}^e_v\Pi_v + \bm{W}^e_s\Pi_{s_i})\} \right),
\end{equation}
where $\Pi_v$ and $\Pi_{s_i}$ are the one-hot embeddings for $v$ and $s_i$, $\bm{W}^e_v \in \mathbb{R}^{d_e \times |\mathcal{V}|}$ and $\bm{W}^e_s \in \mathbb{R}^{d_e \times |\mathcal{SR}| }$ are learnable mapping matrices.

Then, we use a three-layer Transformer decoder to autoregressively generate semantic role sequence (including the verb). 
To prevent the occurrence of semantic role sequence with duplicates, we generate $s_t$ with the highest probability $p(s_t | \mathcal{VSR})$, where $s_t$ is in the input semantic role sequence but hasn't been generated.
\begin{equation}
p(s_t | \mathcal{VSR})  = \text{Transformer}_{\text{dec}} \left( \bm{H}, \bm{W}^e_s\Pi_{\mathcal{S}_{<t}} \right), 
\end{equation}

\noindent\textbf{R-level SSP.}
Since each semantic role $s_i$ has variable number of sub-roles (\ie, $n_i$), we set a constant $n_{\text{max}}$ as the maximum number of sub-roles for each semantic role. We employ the Sinkhorn operation~\cite{mena2018learning} to learn a ``soft" permutation matrix $\bm{P}$. For each proposal $\bm{b}_* \in \mathcal{\hat{B}}$, we encode a feature vector $\bm{\tilde{z}_*}$ by:
\begin{equation}
\begin{aligned}
\bm{\tilde{z}_*} = \text{MLP}_b ([\bm{W}^r_v \bm{f}_*; \bm{W}^r_c \Pi_{c_*}; \text{Pos}(\bm{b}_*)]),
\end{aligned}
\end{equation}
where $\bm{f}_*$ is the detection feature (2048-d); $\Pi_{c_*}$ is GloVe embedding of the region class (300-d); $\text{Pos}(\cdot)$ is a 4-d spatial encoding of $\bm{b}_*$. $\bm{W}^r_x \in \mathbb{R}^{d_v \times v}$ and $\bm{W}^d_c \in \mathbb{R}^{d_c \times |\mathcal{C}|}$ are learnable mapping matrices, $|\mathcal{C}|$ is the size of vocabulary of the detected classes, and $\text{MLP}_b$ is a two-layer MLP to mapping the concatenated feature into $\mathbb{R}^{n_{\text{max}}}$. The position encoding function $\text{Pos}(\cdot)$ encodes the location feature: 
$[\frac{x_{\text{min}}}{W_{\bm{I}}}, \frac{y_{\text{min}}}{H_{\bm{I}}}, \frac{x_{\text{max}}}{W_{\bm{I}}}, \frac{y_{\text{max}}}{H_{\bm{I}}}]$, where $x_{\text{min}}$, $y_{\text{min}}$, $x_{\text{max}}$, $y_{\text{max}}$ are the bounding box coordinates of proposal $\bm{b}_*$; $W_{\bm{I}}$ and $H_{\bm{I}}$ are the width and height of the image $\bm{I}$.

Then, for each proposal set $\mathcal{\hat{B}}_{i} \subset \mathcal{\hat{B}}$, we average-pool all the feature (\ie, $\{\bm{\tilde{z}_*}\}$) of each proposal set, denoted as $\bm{z}_{i}$. And we concatenate all feature representations $\{\bm{z}_{i}\}$ to get a $n_{\text{max}} \times n_{\text{max}}$ matrix $\bm{Z}$. The square matrix $\bm{Z}$ is converted into a ``soft" permutation matrix $\bm{P}$ through the Sinkhorn operator. The operator is K consecutive row-wise and column-wise normalization, as follows:
\begin{equation}
\begin{aligned}
S^0(\bm{Z}) &= \exp(\bm{Z}), \\
S^k(\bm{Z}) &= \mathcal{N}_c(\mathcal{N}_r(S^{k-1}(\bm{Z}))), \\
\bm{P} &= S^K(\bm{Z}),
\end{aligned}
\end{equation}
where $\mathcal{N}_r(\bm{Z}) = \bm{Z} \oslash (\bm{Z} \mathbbm{1}_{n_{\text{max}}} \mathbbm{1}_{n_{\text{max}}}^T)$ and $\mathcal{N}_c(\bm{Z}) = \bm{Z} \oslash (\mathbbm{1}_{n_{\text{max}}} \mathbbm{1}_{n_{\text{max}}}^T \bm{Z})$ are the row-wise and column-wise normalization operations respectively, and $\oslash$ is the element-wise division, $\mathbbm{1}_{n_{\text{max}}}$ is a column vector of ${n_{\text{max}}}$ ones. 

During inference, once $K$ normalizations (we set $K=20$ in our experiments) have been performed, the resulting ``soft'' permutation matrix can be converted into the final permutation matrix via the Hungarian algorithm~\cite{kuhn1955hungarian}.

\subsection{Role-shift Captioning Model}

\noindent\textbf{Adaptive attention for the shifting probability.}
The first LSTM is firstly extended to obtain a sub-role sentinel $\bm{s}^g_t$, which models a component encoding the state of the LSTM at the end of a sub-role. The sentinel is computed as:
\begin{equation}
\begin{aligned}
\label{eq:s_g}
\bm{l}^g_t &= \sigma(\bm{W}_{ig} \bm{x}_t + \bm{W}_{hg} \bm{h}^1_{t-1}) \\
\bm{sr}^g_t &= \bm{l}^g_t\odot\tanh(\bm{m}_t)
\end{aligned}
\end{equation}
where $\bm{W}_{ig} \in \mathbb{R}^{d_l\times d_i}$, $\bm{W}_{hg} \in \mathbb{R}^{d_l\times d_l}$ are learnable weights, $\bm{m}_t \in \mathbb{R}^{d_l}$ is the LSTM cell memory and $\bm{x}_t \in \mathbb{R}^{d_i}$ is the input of the LSTM at time $t$; $\odot$ represents the Hadamard element-wise product and $\sigma$ is the sigmoid function. 

We then compute a compatibility score between the hidden state $\bm{h}^1_t$ and the sentinel vector $\bm{sr}^g_t$ through a single-layer neural network; analogously, we compute a compatibility score between $\bm{h}^1_t$ and the regions in $\bm{r}_t$ by:
\begin{equation}
\begin{aligned}
\hat{\alpha}^g_t &= \bm{w}_{h}^T \tanh(\bm{W}_{sg}\bm{sr}^g_t + \bm{W}_{g} \bm{h}^1_t), \\
\hat{\bm{\alpha}}^r_t &= \bm{w}_{h}^T \tanh(\bm{W}_{sr}\bm{r}_t + (\bm{W}_{g}             \bm{h}^1_t)\mathbbm{1}^T),
\end{aligned}    
\end{equation}
where $\mathbbm{1} \in \mathbb{R}^{n_t}$ is a vector with all elements set to 1, $n_t$ is the number of regions in $\bm{r}_t$, $\bm{w}_h^T$ is a row vector, $\bm{W}_{sg} \in \mathbb{R}^{d_a\times d_l}$, $\bm{W}_{sr} \in \mathbb{R}^{d_a\times d_v}$ and $\bm{w}_h \in \mathbb{R}^{d_a}$ are learnable mapping matrices.

And then we renormalize the attention weight for sub-role sentinel $\bm{sr}^g_t$ over attention weights for the sentinel vector $\bm{sr}^g_t$ and the regions in $\bm{r}_t$:
\begin{equation}
\alpha^g_t = \frac{\exp{\hat{\alpha}^g_t}}{\exp{\hat{\alpha}^g_t} + \sum_{i} \exp{\hat{\bm{\alpha}}^r_{ti}}},
\end{equation}
where $\hat{\bm{\alpha}}^r_{ti}$ indicates the $i$-th element in $\hat{\bm{\alpha}}^r_{t}$.

\noindent\textbf{Adaptive attention for the context feature.}
To further distinguish the textual and visual words, we build an adaptive attention mechanism with a visual sentinel~\cite{lu2017knowing}. The visual sentinel vector models a component which the model can fall back on when it chooses to not attend regions in $\bm{r}_t$. Analogously to Eq.~\eqref{eq:s_g}, it is defined as:
\begin{equation}
\begin{aligned}
\bm{l}^v_t &= \sigma(\bm{W}_{is} \bm{x}_t + \bm{W}_{hs} \bm{h}^1_{t-1}), \\
\bm{sr}^v_t &= \bm{l}^v_t\odot\tanh(\bm{m}_t),
\end{aligned}    
\end{equation}
where $\bm{W}_{is} \in \mathbb{R}^{d_l\times d_i}$ and $\bm{W}_{hs} \in \mathbb{R}^{d_l\times d_l}$ are matrices of learnable weights. Then, the attentive weights are generated over the visual sentinel vector $\bm{sr}^v_t$ and the regions in $\bm{r}_t$: 
\begin{equation}
[\bm{\alpha}^r_t; \alpha^v_t] = \text{softmax}([ \hat{\bm{\alpha}}^r_t; \bm{w}_{h}^T \tanh(\bm{W}_{ss}\bm{sr}^v_t + \bm{W}_{g} \bm{h}^1_t)]),
\end{equation}
where $\bm{W}_{ss} \in \mathbb{R}^{d_a\times d_l}$ is the learnable weights.

\section{Merging Two Semantic Structures} \label{sec: supp4}

The algorithm of merging two semantic structures (\ie, sub-role sequences) is shown in Algorithm~\ref{alg: 1}. Given multiple VSRs, we can continually use this algorithm by regarding the merged semantic structure as the first input structure.

\begin{algorithm}[t] 
	\caption{Merging Algorithm of Semantic Structures}
	\label{alg: 1}
	\hspace*{\algorithmicindent} \noindent\textbf{Input}: Two semantic structures and corresponding sequence of grounded visual regions: ($\mathcal{S}^a$, $\mathcal{R}^a$) and ($\mathcal{S}^b$, $\mathcal{R}^b$). \\
	\hspace*{\algorithmicindent} \noindent\textbf{Output}: The merged semantic structure $\mathcal{S}$ and grounded visual regions $\mathcal{R}$.
	\begin{algorithmic}[1]
		\STATE $\mathcal{R} = \mathcal{R}^a$
		\STATE \textcolor{mygray}{// build a sequence of region sets $\mathcal{R}_\text{same}$, which is in both $\mathcal{R}^a$ and $\mathcal{R}^b$.}
		\FOR{each $\bm{r}^a_i \in \mathcal{R}^a$}
		\IF{$\bm{r}^a_i \in \mathcal{R}^b$}
		\STATE $\mathcal{R}_\text{same}$.append($\bm{r}^a_i$)
		\ENDIF
		\ENDFOR
		\STATE \textcolor{mygray}{// if the rank of the same region sets in $\mathcal{R}^b$ is different from $\mathcal{R}^a$, re-rank those region sets.}
		\STATE $i_\text{same}$ = 0
		\FOR{each $\bm{r}^b_i \in \mathcal{R}^b$}
		\IF{$\bm{r}^b_i \in \mathcal{R}_\text{same}$}
		\STATE $\bm{r}^b_i = \mathcal{R}_\text{same}[i_\text{same}]$
		\STATE $i_\text{same}$ += 1
		\ENDIF
		\ENDFOR
		\STATE \textcolor{mygray}{// insert region sets in $\mathcal{R}^b \setminus R_\text{same}$ into $\mathcal{R}$.}
		\FOR{each $\bm{r}^b_i \in \mathcal{R}^b$}
		\IF{$\bm{r}^b_i \notin \mathcal{R}_\text{same}$}
		\STATE insert $\bm{r}^b_i$ in $\mathcal{R}$ right before $\bm{r}^b_\text{right}$
		\STATE \textcolor{mygray}{// $\bm{r}^b_\text{right}$ is the closest region set in the right of $\bm{r}^b_i$ in $\mathcal{R}^b$, which is also in $\mathcal{R}_\text{same}$.}
		\ENDIF
		\ENDFOR
		\STATE build $\mathcal{S}$ according to $\mathcal{R}$
	\end{algorithmic}
\end{algorithm}

\section{Details of Experimental Settings} \label{sec: supp5}

\noindent\textbf{Parameter Settings.} We use the Adam~\cite{kingma2014adam} optimizer in all our experiments. For the grounded semantic role labeling model, we initiate the learning rate to $1\times10^{-5}$, which decreases by a factor of 0.5 for every 3 epochs. To train the S-level SSP and R-level SSP, the learning rate is set to $1\times10^{-4}$ and decreases by a factor of 0.6 for every 3 epochs. And the max training epoch is set to 20 for the models above.
For the role-shift captioning model, the batch size is set to 100. The learning rate is $5\times10^{-4}$ for XE training and $5\times10^{-5}$ for the RL training, decreasing by a factor of 0.8 every epoch. The hidden size of both two LSTMs is set to 512. In the training stage, we apply early stopping according to the CIDEr-D score in the validation dataset. In the inference stage, we employ the beam search strategy with a beam size of 5. 

\noindent\textbf{Details of Training and Test.} Due to the constraint of COCO/Flickr30k Entities, there are many captions containing nouns without region annotation. Thus, we followed~\cite{cornia2019show} to fill the missing regions with most probable detections of the image in the training of role-shift caption model and drop these captions in validation and test stages. And those are also dropped in other models' training and test stages.

\section{Transformer vs. Sinkhorn Network in the S-level SSP.} \label{sec: supp6}

\begin{table}[t]
	\small
	\begin{center}
		\scalebox{0.95}{
			\begin{tabular}{| l | c | c | c c c c c| }
				\hline
				& Proposal & Model & B4 & M & R & C & S \\
				\hline	\parbox[t]{2mm}{\multirow{4}{*}{\rotatebox[origin=c]{90}{COCO}}}
				& \multirow{2}{*}{GSRL} & SN & 15.5 & 23.0 & 46.5 & 159.3 & 35.1 \\ 
				& & TF & \textbf{16.0} & \textbf{23.2} & \textbf{47.1} & \textbf{162.8} & \textbf{35.7} \\
				\cline{2-8}
				& \multirow{2}{*}{GT} & SN & 22.3 & 27.6 & 54.2 & 227.9 & 48.1 \\
				& & TF & \textbf{23.1} & \textbf{28.0} & \textbf{55.6} & \textbf{235.1} & \textbf{48.9} \\
				\hline	\parbox[t]{2mm}{\multirow{4}{*}{\rotatebox[origin=c]{90}{Flickr30K}}} & \multirow{2}{*}{GSRL} & SN & 7.6 & 14.5 & 32.1 & 69.0 & 17.8 \\
				& & TF & \textbf{7.9} & \textbf{14.7} & \textbf{32.6} & \textbf{71.6} & \textbf{18.2} \\
				\cline{2-8}
				& \multirow{2}{*}{GT} & SN & 9.6 & 17.3 & 35.4 & 86.9 & 21.2 \\
				& & TF & \textbf{10.7} & \textbf{18.0} & \textbf{37.1} & \textbf{97.5} & \textbf{21.9} \\
				\hline
			\end{tabular}
		} 
	\end{center}
	\caption[]{Performance comparisons between Transformer (TF) and Sinkhorn Network (SN) in S-level SSP on dataset COCO Entities and Flickr30K Entities.}
	\label{tab:TFvsSN}
\end{table}

\noindent\textbf{Settings.} To sort the sequence of roles from the given control signal, Sinkhorn network is another alternative network. To further compare the Transformer and Sinkhorn network in the S-level SSP, we design a strong baseline by replacing the Transformer to Sinkhorn network. The results on COCO Entities and Flickr30K Entities are reported in Table~\ref{tab:TFvsSN}.

\noindent\textbf{Results.} From Table~\ref{tab:TFvsSN}, we can observe that the model with Transformer can achieve better performance than the model with Sinkhorn network in all proposal settings (GSRL detected proposals or ground truth proposals) and evaluation metrics on both COCO Entities and Flickr30K Entities benchmarks. This may because that the Transformer can better encode the dependency on previous outputs (semantic roles). Thus, we use Transformer for our S-level SSP.

\end{normalsize}

\end{document}